\documentclass[10pt,journal,compsoc]{IEEEtran}
\usepackage{graphicx}
\usepackage[hidelinks]{hyperref}
\usepackage{amsmath}
\usepackage{algorithm}
\usepackage{enumitem}
\usepackage{float}
\usepackage{adjustbox}
\usepackage{algpseudocode}
\usepackage{graphicx}
\usepackage{varioref}
\usepackage{longtable,pdflscape,booktabs}
\usepackage{array}
\usepackage{booktabs}
\usepackage{multirow}
\usepackage{subcaption}
\usepackage{array,arydshln}
\usepackage{amssymb}
\usepackage{hhline}
\usepackage{times,rotating}
\usepackage{comment}
\usepackage{caption}
\usepackage{enumitem}
\usepackage{tabularx}
\setlist[itemize]{noitemsep}

\begin{document}

\title{Explainable Artificial Intelligence for Autonomous Driving: A Comprehensive Overview and Field Guide for Future Research Directions}

\author{Shahin Atakishiyev,
    Mohammad Salameh,
    Hengshuai Yao,
    Randy Goebel 
\IEEEcompsocitemizethanks{
\IEEEcompsocthanksitem Shahin Atakishiyev and Randy Goebel are with the Department of Computing Science, University of Alberta. Hengshuai Yao is with Sony AI (The work was done while he was with Huawei Technologies Canada Co., Ltd).
Mohammad Salameh is with Huawei Technologies Canada Co., Ltd. Primary contact: \texttt{shahin.atakishiyev@ualberta.ca}

}
\thanks{Manuscript has been submitted for possible publication.}}

\markboth{}%
{Shell \MakeLowercase{\textit{et al.}}: Explainable Artificial Intelligence for Autonomous Driving: A Comprehensive Overview and Field Guide for Future Research Directions}

\IEEEtitleabstractindextext{%
\begin{abstract}
Autonomous driving has achieved significant milestones in research and development over the last two decades. There is increasing interest in the field as the deployment of autonomous vehicles (AVs) promises safer and more ecologically friendly transportation systems. With the rapid progress in computationally powerful artificial intelligence (AI) techniques, AVs can sense their environment with high precision, make safe real-time decisions, and operate reliably without human intervention. However, intelligent decision-making in such vehicles is not generally understandable by humans in the current state of the art, and such deficiency hinders this technology from being socially acceptable. Hence, aside from making safe real-time decisions, AVs must also explain their AI-guided decision-making process in order to be regulatory compliant across many jurisdictions. Our study sheds comprehensive light on the development of explainable artificial intelligence (XAI) approaches for AVs. In particular, we make the following contributions. First, we provide a thorough overview of the state-of-the-art and emerging approaches for XAI-based autonomous driving. We then propose a conceptual framework that considers the essential elements for explainable end-to-end autonomous driving. Finally, we present XAI-based prospective directions and emerging paradigms for future directions that hold promise for enhancing transparency, trustworthiness, and societal acceptance of AVs.
\end{abstract}

}

\maketitle

\IEEEdisplaynontitleabstractindextext

\IEEEpeerreviewmaketitle

\ifCLASSOPTIONcompsoc
\IEEEraisesectionheading{\section{Introduction}\label{sec:introduction}}
\else
\section{Introduction}
\label{sec:introduction}
\fi

\begin{figure}[htp]
    \centering
    \vspace{-0.15cm}\includegraphics[width=9 cm]{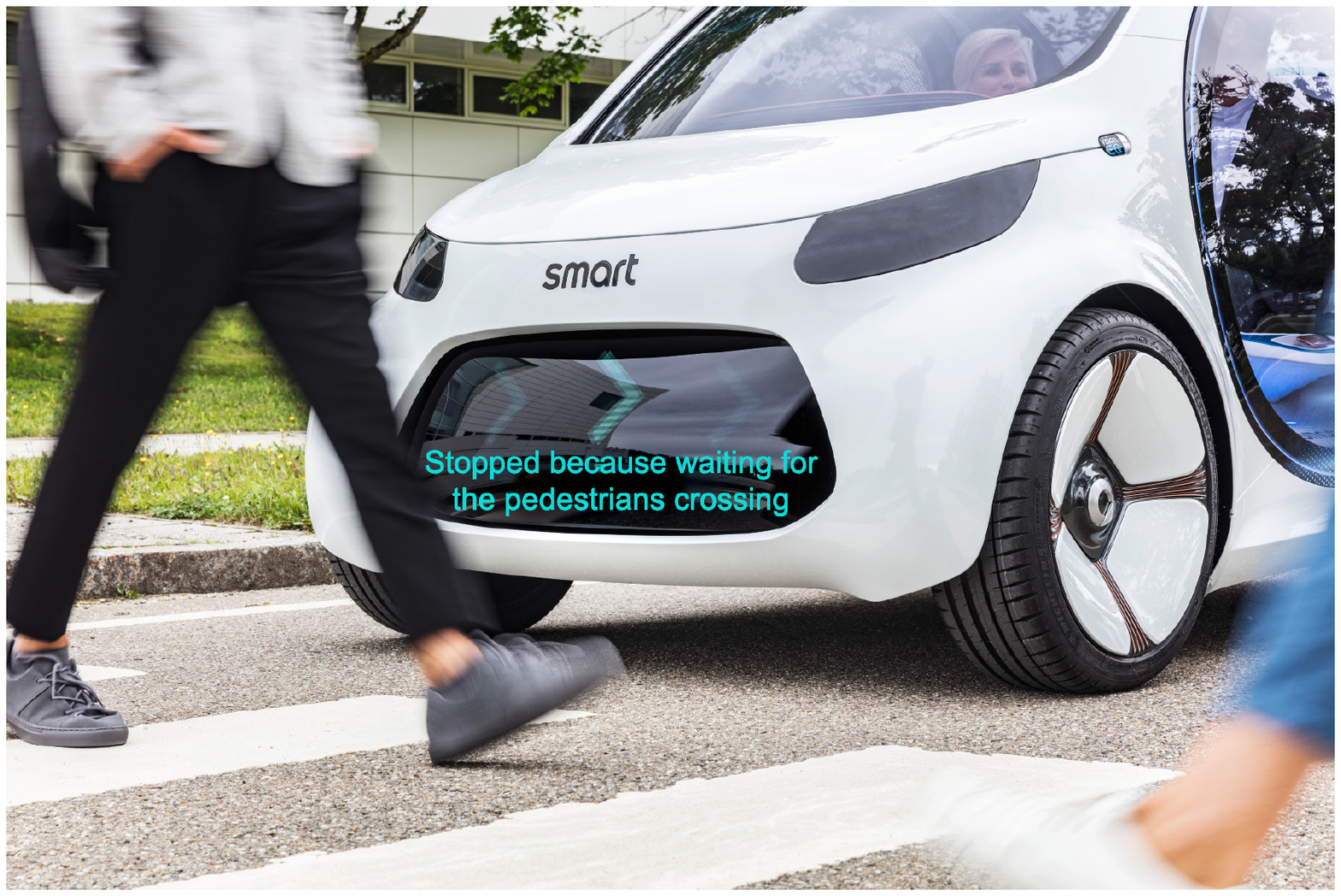}
    \caption {A canonical example of explainable AI in autonomous driving: An autonomous vehicle provides a live natural language explanation of its real-time decision to bystanders. The image has been adapted and modified from the original source: \cite{daimler2017}.}
    \label{fig:car_explainability}
\end{figure}

\IEEEPARstart{A} survey of the American National Highway Traffic Safety Administration (NHTSA) reports that nearly 94\% of road accidents are due to human errors \cite{singh2015critical}. Such a lack of rule obedience and improper road culture have, therefore, motivated officials, manufacturers, and legislators to make substantial improvements in transportation systems. In this sense, there are growing research and development attempts to enhance safety and automation capability of AVs with the goal of preventing traffic accidents, and creating a better road infrastructure. Intel’s report on the projected benefits of AVs estimates that deployment of this technology on roads will result in a reduction of 250 million hours of users’ commuting time per year and save more than half a million lives from 2035 to 2045, just in the USA \cite{lanctot2017accelerating}.\\
While the potential impact and benefits of AVs in everyday life are promising, there is a major societal concern about functional safety of such vehicles. This issue, as a major drawback, originates mainly from reports of recent traffic accidents with the presence
of AVs, primarily owing to their ``black-box" decision-making \cite{adadi2018peeking, stanton2019models, yurtsever2020survey,board2020collision}. As AI approaches provide the foundation for real-time driving actions, there is an inherent need and expectation from consumers, general society, and regulatory bodies that AI-based action decisions of AVs should be explainable to build confidence in these vehicles \cite{adadi2018peeking, ali2023explainable, dong2023did, saeed2023explainable} (e.g., Figure \ref{fig:car_explainability}). \\
In this survey, we present a comprehensive overview of state-of-the-art investigations on the explainability of autonomous driving.

\hspace{-0.62cm}
Through extensive analyses, we first show the background information on the need for the emergence of explanations for AVs. Furthermore, we fill the gap in the current literature by providing a structured and comprehensive review of state-of-the-art and emerging XAI approaches for autonomous driving and present a road map for future directions. More specifically, we discuss the following research questions in depth:

\begin{enumerate}[leftmargin=*]
\item Why is there a need for XAI in AVs technology?
\item What are the current trends and emerging AI technologies for explainable autonomous driving?
\item What are promising future XAI directions toward trustworthy,

 \begin{figure*}[htp]
 \centering
    \includegraphics[width=17.5 cm]{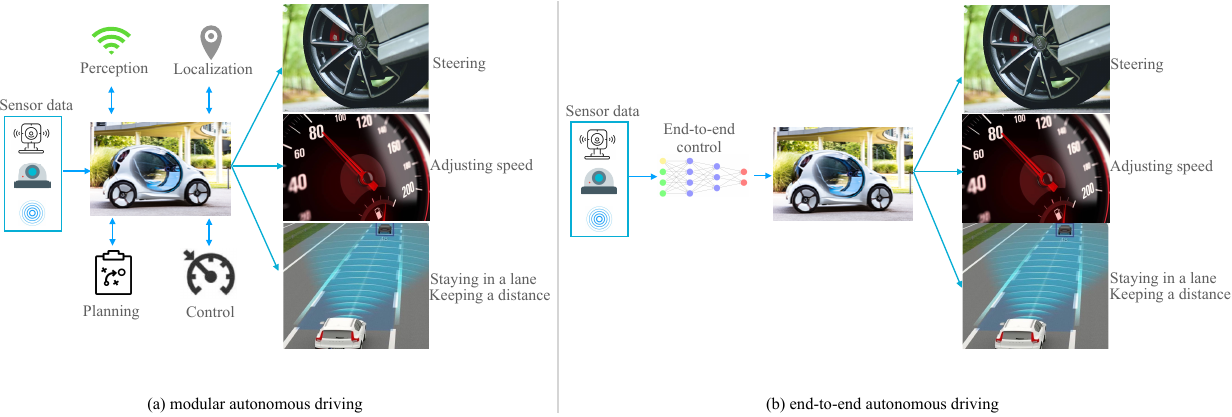}
    \caption{Modular vs. end-to-end autonomous driving. In the modular pipeline, the described operations are carried out subsequently to produce control commands while end-to-end driving directly inputs raw sensor data and produces control commands as a unified task.}
    \label{fig:auto_decision_making}
\end{figure*}
\hspace{-0.60cm}
responsible, regulatory compliant, and publicly acceptable AVs?
\end{enumerate}
With these questions in mind, our paper makes the following contributions:
\begin{itemize}[leftmargin=*]

\item We describe cross-disciplinary perspectives necessitating explainability in autonomous driving;
\item We provide a survey of state-of-the-art XAI-based investigations for autonomous driving;
 \item We present a conceptual design framework for explainable end-to-end autonomous driving;
\item We propose future research directions on promising XAI approaches for autonomous driving.
\end{itemize}
 The rest of the article consists of six sections. Section 2 provides background information and the factors triggering the need for the emergence of XAI in autonomous driving. Section 3 covers the concept of explainability for AVs by analyzing 1) cross-disciplinary perspectives necessitating explanations, 2) the role of various types of explanations for diverse explanation recipients, and 3) construction methodologies for explanations. Section 4  provides a comprehensive survey of studies on XAI-based autonomous driving. Motivated by current limitations and trends delineated in these works, Section 5 presents a general design framework for explainable autonomous driving and shows key components of such a framework. Finally, Section 6 outlines potential challenges and a road map toward safety and explainability of next-generation AVs, as future directions, and Section 7 concludes the paper with an overall summary.

 \section{Background} 
\subsection{Autonomous driving at a glance}
AVs, also referred to as self-driving vehicles, are intelligent vehicles equipped with advanced sensors, cameras, RADAR, LIDAR, GPS, and sophisticated learning algorithms that enable them to navigate and operate without human intervention \cite{badue2021self}. To discern, identify, and distinguish the objects in their operational surroundings, these vehicles fuse information from a variety of sensors that help make real-time driving decisions \cite{campbell2018sensor, yeong2021sensor}. The history of contemporary AVs goes back to 1988, when ALVINN (Autonomous Land Vehicle In a Neural Network), the first neural network-powered self-driving vehicle taking camera images with a laser range finder, was able to produce control commands for the road-following task \cite{pomerleau1988alvinn}. Current AVs deployed on road networks have different levels of automation based on their in-vehicle technologies and intelligent capabilities. SAE International (previously known as the Society of Automotive Engineers) has defined six levels of autonomous driving \cite{Shuttleworth}: Level 0 - No automation (a human driver is responsible for all critical driving tasks); Level 1 - Driving assistance (a vehicle has automated driving support such as acceleration/braking or steering, but the driver is responsible for all other possible driving operations); Level 2 - Partial automation (Advanced Driving Assistance Systems (ADAS) operations such as steering and acceleration/braking are available in this level); Level 3 - Conditional automation (a vehicle has more advanced features such as object/obstacle detection and can carry out the majority of driving operations); Level 4 - High automation (a vehicle can fulfill all possible driving operations in a geofenced area); and Level 5 - Full automation (a vehicle can perform all driving operations in any likely scenario, and no human intervention is required). \\
There are two main approaches to building autonomous driving systems in terms of their AI-based learning architecture: modular and end-to-end pipelines \cite{yurtsever2020survey, chen2023end}. The modular pipeline consists of four primary and interconnected modules categorized as \textit{perception}, \textit{localization}, \textit{planning}, and \textit{control} (Figure \ref{fig:auto_decision_making}, a). The modular pipeline leverages various sensor suites and algorithms for each module. While being comprised of standalone components makes the modular system more explainable and debuggable, such an architecture propagates errors to the next component, and thus, the overall pipeline error becomes cumulative \cite{chen2023end, hu2023planning,  araluce2024leveraging}. \\
In contrast to a modular pipeline, end-to-end autonomous driving has recently emerged as a paradigm shift in the design and development of AVs. End-to-end autonomous driving takes the raw sensor data as visual input and yields a control command for the vehicle (Figure \ref{fig:auto_decision_making}, b) \cite{chen2023end, tampuu2020survey, hu2022st}. Particularly, recent breakthroughs in deep learning and computer vision algorithms, and the availability of rich sensor devices along with enhanced safety benefits have been the primary reasons for automotive researchers to leverage the end-to-end learning approach. The advantage of an end-to-end pipeline over its counterpart is that it directly produces driving actions by unifying perception, localization, planning, and control as one combined machine learning (ML) task. Furthermore, computational efficiency is improved via shared backbones in end-to-end learning, and in this way, potential information loss in intermediate layers is also avoided \cite{chen2023end, jing2022inaction}.

\subsection{Fundamental issues}
AI approaches, which are currently predominated by deep learning algorithms, have brought considerable improvements to many essential components of autonomous driving technology, including advances in perception, object detection, and planning. As the AI-powered driving systems of vehicles advance, the number of AVs deployed to road networks has proliferated significantly in many developed European countries, the US, and Canada over the last decade \cite{muller2019comparing}. However, the aforementioned road accidents involving such cars have caused public skepticism, and many studies have attempted to underscore the current limitations and issues with the design, development, and deployment of AVs on roads. For example, Fleetwood \cite{fleetwood2017public} has investigated public health and ethical issues arising with the use of autonomous driving. Their study provides an in-depth analysis of the health issues, especially with the Trolley problem examples \cite{foot1967problem, jarvis1985trolley} (hitting a pedestrian on an icy road or a parked car; driving and hitting five people or changing the direction of the steering wheel and hitting an individual, etc.). Some studies have directly focused on the concept of ethical crashing (i.e., if crashing is inevitable, how to crash?) and the Trolley problem mentioned above. For instance, the Moral Machine experiment \cite{awad2018moral}, a well-known and hotly debated experiment, investigates a general community's preferences on applied Trolley problems (inevitable accident scenarios with binary outcomes) and states that ``these preferences can contribute
to developing global, socially acceptable principles for machine ethics.'' However, further discussion on this issue condemns this opinion and draws attention to the lack of safety principles \cite{lundgren2020safety}, which force deeper consideration of such dilemmas \cite{harris2020immoral}. 
Burton et al. \cite{burton2020mind} have identified three open problems in the state-of-the-art development of autonomous systems. The first one is the \textit{semantic gap} that emerges when a thorough specification of the system is not provided to manufacturers and designers. Another identified issue is the \textit{responsibility gap}, which arises when an accident happens and the responsibility of either an autonomous system or a human is the cause of this accident remains unresolved. Finally, there is the question of who is responsible for compensating the injured during an accident, which precipitates the third issue: the \textit{liability gap}. That study also shows that the core of these issues is associated with domain complexity, system complexity, and transferring more decision-making functions from humans to autonomous systems. Further studies include the outcomes of autonomous driving technology on public health in an urban area \cite{sohrabi2020impacts}, and ethical dilemmas with AVs \cite{martinho2021ethical}. Overall, the key findings from these studies necessitate an understanding of the causes of these issues and intrinsically give the stakeholders the right to ask ``why'' questions.
\subsection{Regulations and standards}
The issues and growing concerns caused by AI systems create the need to scrutinize the regulation of this technology. As a result, public institutions have initiated the development of regulatory frameworks to monitor the activities of data-driven systems at both a country level and internationally. The focal points of these regulations are mainly to protect the stakeholders’ rights and ensure they have control over their data. For example, the General Data Protection Regulation (GDPR) of the European Union (EU) initiated guidelines to promote the ``right of an explanation'' principle for users, enacted in 2016 and taking effect in May 2018 \cite{regulation2016}. Moreover, the EU has a specially defined strategy on Guidelines of Trustworthy AI that has seven essential requirements, namely 1) human agency, 2) technical robustness and safety, 3) privacy and data governance, 4) transparency, 5) accountability, 6) diversity, non-discrimination, and fairness, and 7) societal and environmental well-being; these principles are all to be applied in AI-based product research and development \cite{eustrategy2019}. \\
Various organizations have recently proposed guidelines on the regulation of AVs to monitor their compliance with law enforcement. NACTO's (National Association of City
Transportation Officials) statement on automated vehicles proposes nine principles to shape a policy on regulation of future generation AVs \cite{nacto2016}. NHTSA of the US Department of Transportation has a specific federal guideline on automated vehicle policy to improve traffic safety \cite{national2016federal}. In March 2022, NHTSA announced that automobile manufacturers would no longer have to equip fully autonomous cars with manual control elements, such as a steering wheel and braking pedals in the USA \cite{NHTSA2022}. Canada \cite{guideleinescanada}, Germany \cite{germany2021}, UK  \cite{UK2021}, Australia \cite{austroads2017guidelines}, and Japan \cite{japan2017} have also recently launched their regulations on autonomous driving technology.\\
While the regulations have been set out to ensure legislative norms and user demands are met, some standards provide specifications to achieve a high safety level, quality assurance, efficiency, and environmentally friendly transportation systems. The International Organization for Standardization (ISO) has adopted several standards to define the relevant issues on automated driving. Examples include the ISO 21448 \cite{ISO21448}, which specifies situation awareness standards to maintain operational safety under the ``Safety of the Intended Functionality,'' and the ISO 26262 \cite{ISO26262} standard defined for the safety of electrical and electronic systems in production passenger vehicles, entitled as ``Road vehicles – Functional safety.'' Thorough documentation on the details of legislation, regulation, and standardization of autonomous cars can be viewed in \cite{apexdocs}.

\section{Explanations in autonomous driving}
\subsection{The need for explanations in AVs}
The need for explanations in autonomous driving arises from fundamental issues, established regulations and standards covered in previous subsections, and cross-disciplinary views and opinions of society. At the highest level, the necessity of explanations for AVs can be summarized in terms of four perspectives: 
\begin{itemize}[leftmargin=*]
    \item \textit{Psychological} perspective: Traffic accidents and safety concerns remain the main cause of the need for XAI in autonomous driving from a psychological point of view \cite{omeiza2021explanations}.
    \item \textit{Sociotechnical} perspective: The design, development, and deployment of AVs should be human-centered, reflecting the target audience’s needs, and taking their prior opinions and expectations into account \cite{ehsan2020human, dhanorkar2021needs}.
    \item \textit{Philosophical} perspective: Explaining AI decisions can provide descriptive information about the causal history of actions performed, particularly in critical situations \cite{lewis1986causal, miller2019explanation, pearl2009causality}.
    \item \textit{Legal} perspective: It considers all the above-mentioned factors and incorporates them into general regulatory compliance principles for AVs. A notable example is GDPR's requirements on explanation provision for end users \cite{regulation2016}. 
\end{itemize} 
Overall, we can conclude that the explainability of autonomous driving systems is an expectation and a requirement from a multidisciplinary point of view.
\subsection{Potential benefits of explanations for AVs}
Considering  these multi-dimensional perspectives, explainable autonomous driving can bring the following benefits to the stakeholders: 
\begin{itemize}[leftmargin=*]
    \item \textit{Human-Centered Design}:
Getting the end users' inputs, opinions, and anticipations on the design and development of the semi or fully AVs can help with the acceptance of this technology by the general community  \cite{design2016vision}. 
\item \textit{Trustworthiness}: Algorithmic assurance can build trust in human-autonomous system relationships \cite{israelsen2019dave}.

\item \textit{Traceability}: Explainable intelligent driving systems can help forensic analysts and system auditors understand the entire decision-making process of an autonomous car during the journey via a post-trip analysis. 
\item \textit{Transparency and accountability}: Explanations can help achieve accountability, which can resolve the potential liability and responsibility gaps in foreseeable post-accident investigations with the involvement of autonomous cars as described by Burton et al. \cite{burton2020mind}. For example, Mercedes-Benz has recently taken a promising step forward and announced that the corporation will take legal responsibility for any accidents that their self-driving systems are engaged in \cite{Mercedes2022}. Mercedes's declaration of legal culpability is a significant milestone toward the accountability of AVs technology.
\end{itemize}
\subsection{Explanation recipients in AVs}
\begin{figure}[htp!]
    \centering
    \includegraphics[width=9 cm]{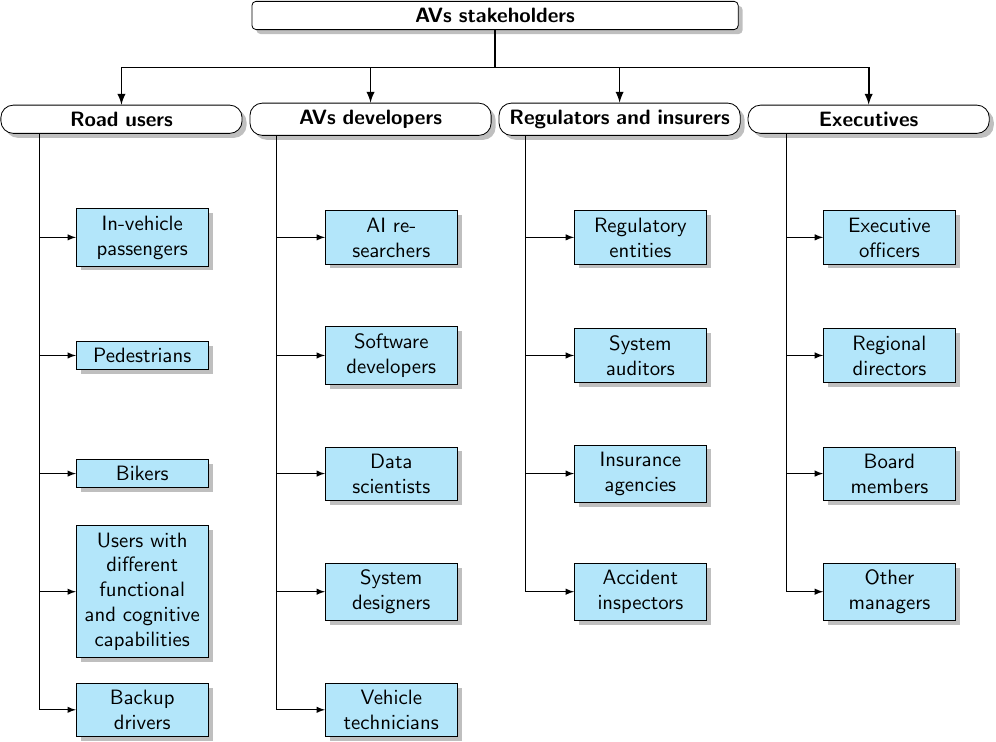}
    \caption{ Taxonomy of the stakeholders in autonomous driving.}
    \label{fig:stakeholders}
\end{figure}

The details, types, and delivery of explanations vary in accordance with users’ identities, technical background knowledge in autonomous driving, and their various functional and cognitive abilities \cite{omeiza2021explanations, arfini2023design}. For instance,  a user having little technical expertise on how AVs operate may be satisfied with a simple explanation of a relevant decision/outcome. However,  an autonomous systems engineer will need more informative explanations to understand the current functionalities of the car, with the motivation to appropriately ``debug'' the existing driving system as required. Therefore, the use of domain knowledge and expertise of the explainee is essential to provide pertinent, sufficiently informative, and intelligible explanations \cite{langley2019varieties, mittelstadt2019explaining}. Motivated by a target audience definition of \cite{omeiza2021explanations} and  \cite{arrieta2020explainable}, we can distinguish \textit{four} groups of the stakeholders in autonomous driving, namely Group 1 - Road users, Group 2 - AVs developers, Group 3 - Regulators and insurers, and Group 4 - Executive management of automobile companies. Figure \ref{fig:stakeholders}
provides the identity of such stakeholders and their positions in the relevant classification.

\subsection{How to deliver explanations in AVs?}
As explainees are classified based on their domain knowledge and needs, explanations and their design and evaluation techniques also vary depending on the context and knowledge of the category of explainees. In fact, explanation construction is one of the major challenges in current XAI research. Zablocki et al. \cite{zablocki2021explainability} define four ``W'' questions in XAI-based autonomous driving: 1) Who needs explanations? 2) Why are explanations needed? 3) What kind of explanations can be generated? and 4) When should explanations be delivered? In general, explanations in AI can be distinguished

\hspace{-0.65cm}
based on their \textit{derivation category} and  \textit{classification}. Some of
the early practical studies have applied explanations to automated collaborative filtering systems \cite{herlocker2000explaining} and knowledge-intensive case-based reasoning \cite{roth2004explanations}. Another empirical approach attempted to derive explanations based on some intelligibility types \cite{lim2009assessing} and used ``why,'' ``why not,'' ``what if,'' and ``how to'' type explanations for causality filtering. Furthermore, Liao et al. \cite{liao2020questioning} have interviewed twenty user-interface and design practitioners working in different areas of AI to understand users’ explanatory requirements. By doing so, they have attempted to find the gaps in the interviewers’ products and developed \textit {a question bank}: the authors represent users’ needs as questions so that users may potentially ask about the outcomes produced by an AI system. Overall, the stakeholder needs-based explanation design can be viewed as one of the promising approaches.\\
Another popular approach to producing explanations is based on using psychological tools from \textit{formal theories}, according to the literature review of \cite{wang2019designing}. Depending on the context and addressee, both explanation derivation methods confirm their usefulness. These explanation generation approaches can find alignment in their application in autonomous driving; since autonomous driving involves people with diverse backgrounds in society, relevant XAI design needs inherent adjustments to the context problem. \\ 
Except for \textit{informational content},
the effective communication of explanations is also a key factor for good human-machine teaming. In general, the conveyance of explanations to end users is realized through a user interface (UX) or a human-machine interface (HMI) \cite{naujoks2018use}. For instance, an HMI may be an interface to alert the human driver to take over the control of an vehicle in an emergent situation. Other potential examples are display monitors, sound, light signals, and vibrotactile technology that explain the vehicle's decision-making intentions and bring situation awareness for people in the loop, as shown in Schneider et al.'s work \cite{schneider2021increasing}. 

\section{XAI for autonomous driving: A survey}
\subsection{Previous surveys on XAI for AVs}
There exist reviews on XAI for AVs, which provide valuable insights from various perspectives. These studies expose a variety of approaches, from a universal look to an algorithmic point of view. In this sense, the first noteworthy review on XAI for AVs is 

\begin{table*}[t!]
\normalsize
\caption{Studies on visual explanations for AVs}
\captionsetup[table]{position=above}
\centering
\resizebox{\textwidth}{!}{%
    \begin{tabular}{c p{6cm} p{6cm} p{6cm} p{6cm} c}
        \specialrule{.2em}{.1em}{.1em}
        \multirow{3}{*}{\textbf{Study}} & \multicolumn{3}{c|}{\textbf{}} & \multirow{3}{*}{\textbf{Target audience}} \\ 
        & \textbf{Task} & \textbf{Algorithms/Methods} & \textbf{Delivery format} & \\ \specialrule{.2em}{.5em}{.5em}
        Bojarski  et  al., ~ \cite{bojarski2016visualbackprop}, 2016 & Pixel-based explanations of CNN predictions  & CNN & Visual & AVs developers  \\
        \hline
        Kim and Canny~ \cite{kim2017interpretable}, 2017   & Explaining behavior of a vehicle controller using heat maps  &  CNN, LSTM & Visual& AVs developers  \\
        \hline
        Kim et al.,~ \cite{kim2018textual}, 2018  & Generating textual explanations on a vehicle's control commands  & CNN, S2VT, LSTM & Visual and Textual & All groups\\
        
        \hline
        Hofmarcher et al.,~ \cite{hofmarcher2019visual}, 2019 & Visual scene understanding using semantic segmentation  & Enet, SqueezeNet 1.1, ELU & Visual & AVs developers\\
        
        \hline
        Zeng et al.,~\cite{zeng2019end}, 2019    & End-to-end interpretable neural motion planner  & FaF, IntentNet & Visual & AVs developers\\

        \hline
        Hu et al.,~\cite{hu2019multi}, 2019    & Interpretable multi-modal probabilistic prediction for autonomous driving  & CVAE, Dynamic time warping, LSTM & Visual & AVs developers\\
        
        \hline
        Xu et al., ~\cite{xu2020explainable}, 2020   & Explaining object-induced action decisions for autonomous vehicles  & Faster R-CNN & Visual & All groups \\
        \hline
        Kim et al., ~\cite{kim2020advisable}, 2020   & Advisable learning for self-driving vehicles
by internalizing observation-to-action rules & Mask~R-CNN, LSTM & Visual and Textual & All groups \\
         \hline
    Li et al.,~\cite{li2020make}, 2021    & Risk object identification via causal inference & InceptionResnet-V2, Mask R-CNN, Deep SORT, RoIAlign & Textual & All groups \\ \hline
     
    Casas et al.,~\cite{casas2021mp3}, 2021    & End-to-end model for mapless autonomous driving & CoordConv & Visual and Textual & All groups \\ 
        
        \hline
        Kim et al., ~\cite{kim2021toward}, 2021    & Explainable and advisable model for self-driving cars  & DeepLab v3, Mask~R-CNN, LSTM & Textual & All groups  \\ \hline
         Wang et al., ~\cite{wang2021human}, 2021    & Enhancing automated driving with human
foresight  & Gaze-based vehicle reference & Visual & Road users \\
\hline
        Chitta et al.,~ \cite{chitta2021neat}, 2021    & Interpretable neural attention fields for end-to-end autonomous driving  & ResNet, MLP& Visual & AVs developers \\ 

   \hline
        Dong et al.,~ \cite{dong2021image}, 2021    & Explainable autonomous driving via an image transformer & ResNet-50, Mobilenet-v2, multi-head self-attention & Textual & All groups \\ 
            \hline
            
        Hanna et al., \cite{hanna2021interpretable}, 2021    & Interpretable goal recognition in the presence of occluded factors for autonomous vehicles  & Goal and Occluded Factor Inference, Monte Carlo Tree Search & Visual & AVs developers \\
            \hline
        Mankodiya ~et~al., \cite{mankodiya2021xai}, 2021    & XAI for trust management in autonomous vehicles & Random Forest, Decision Tree, AdaBoost & Visual & AVs developers \\

        \hline
        Madhav and Tyagi, \cite{madhav2022explainable}, 2022    & Explainable navigational intelligence for trustworthy autonomous driving & Grad-CAM, Lime & Visual & AVs developers \\
   \hline
        Jing et al., \cite{jing2022inaction}, 2022    & Interpretable action decision making for autonomous driving & Faster R-CNN & Visual and Textual & All groups \\

    \hline
        Jacob et al., \cite{jacob2022steex}, 2022    & Region-targeted counterfactual explanations & GANs & Visual & AVs developers \\
\hline
        Zhang et al., \cite{zhang2022attention}, 2022    & Interrelation modeling for explainable automated driving & Faster R-CNN, ResNet-50 & Visual & AVs developers \\
   \hline
        Kolekar et al., \cite{kolekar2022explainable}, 2022 & Traffic scene understanding via U-Net and Grad-CAM & U-Net, GradCam & Visual  & AVs developers \\
   \hline

        Zemni et al., \cite{zemni2023octet}, 2023    & Object-aware counterfactual explanations & BlobGAN & Visual & AVs developers \\
   \hline
        Itkina and Kochenderfer \cite{itkina2023interpretable}, 2023 & Trajectory prediction via interpretable self-aware neural networks & PostNet & Visual  & AVs developers \\

   \hline
        Feng et al., \cite{feng2023nle}, 2023    & Natural language explanations via semantic scene understanding & DeepLabV3 & Visual and textual & All groups \\
   \hline
        Hu et al., \cite{hu2023holistic}, 2023    & Interpretable trajectory prediction and decision-making of AVs & LaneGCNN, ResNet & Visual & AVs developers \\
   \hline
        Dong et al., \cite{dong2023did}, 2023    & Describing traffic scenes in 
 natural language via attention-based transformer & CNN, LSTM, Transformer & Visual and textual & All groups \\

  \hline
        Atakishiyev et al., \cite{atakishiyev2023explaining}, 2023    & Explaining autonomous driving actions with visual question answering & VGG-19, LSTM, DDPG & Textual & All groups \\

        \hline
        Echterhoff et al., \cite{echterhoff2024driving}, 2024 & Leveraging concept
bottlenecks as visual features for predicting control command and explanations of vehicle and human behavior & Longformer, GPT 3.5 & Visual and textual & All groups \\
 \hline
        Feng and Sun \cite{feng2024polarpoint}, 2024 & Interpreting self-driving decisions and improving safety by paying more attention to the regions that are near the ego vehicle & Multilayer Perceptron, Trajectory-guided Control Prediction & Visual & AVs developers \\

  \hline
        Araluce et al., \cite{araluce2024leveraging}, 2024 & Using driver attention for an end-to-end explainable decision-making from frontal driving images & ARAGAN, MobileNetV2 & Visual & AVs developers \\
        \specialrule{.2em}{.1em}{.1em}
    \end{tabular}%
}

\label{tab:vision_based_explanations}
\end{table*}
\hspace{-0.62cm}
Omeiza et al.'s work \cite{omeiza2021explanations}. They study the need for/role of explanations for autonomous driving, and focus on legal requirements, standards, and consumer expectations for the design and development of explainable autonomous driving systems.  This provides their basis to present a conceptual XAI framework for modular autonomous driving. In further work, Zablocki et al. \cite{zablocki2021explainability} present a detailed overview of end-to-end vision-based autonomous driving systems and describe explainability hurdles for AVs from an ML perspective. Finally, in very recent work, Kuznietsov et al. \cite{kuznietsov2024explainable} present a systematic review of XAI techniques for modular and end-to-end autonomous driving by focusing on how such techniques improve safety and user trust. In this regard, they propose the SafeX framework for modular autonomous driving by integrating user interface, safety, and explainability. \\
Our study, as complement, extends the coverage of this previous work in three essential dimensions. First, all three previous surveys specifically focus on \textit{form} and \textit{content} of explanations; however, \textit{time granularity} of explanations has not been investigated. As AVs are real-time decision-making systems, it is crucial to know how explanations must be delivered from the timing perspective. Furthermore, attention-based transformers, large language models, and vision-language models are now at the forefront of AI applied to AVs technologies, and such approaches have not been explored in the aforementioned surveys. Finally, a classification of XAI approaches applied to AVs from an algorithmic/methodological perspective is also a noteworthy nuance missing in these studies. Consequently, our paper extends the above-mentioned reviews by (1) analyzing the temporal sensitivity of explanations, (2) providing a perspective on emerging XAI paradigms, and (3) presenting an methodological taxonomy of XAI methods for AVs. 

\subsection{Structure of our survey}
Our survey presents a comprehensive overview of XAI methods for AVs. In particular, we present a classification of approaches in terms of vision, reinforcement learning, imitation learning, decision trees, logic, user study, and the most recent paradigm

\begin{figure*}[htp!]
    \centering
    \includegraphics[width=1\textwidth]{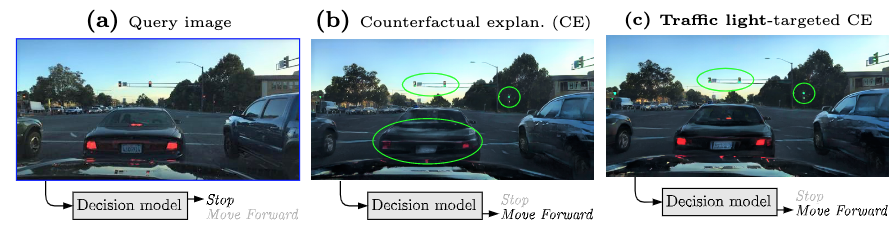}
    \caption{An example of a counterfactual explanation generated by STEEX. Graphics credit: \cite{jacob2022steex}.}
    \label{fig:counterfactual_explanations}
\end{figure*}\hspace{-0.62cm}
shift - large language and vision-language-based explanations for AVs. By adopting insights from the recent industrial trends, emerging AI technologies, and general regulatory-compliance principles, we further present a conceptual framework for explainable end-to-end autonomous driving and describe its essential elements. Finally, we present a set of promising XAI approaches by describing the missing pieces in the state of the art and present potential solutions with the goal of bridging the gaps and achieving transparency and social acceptance in the next-generation AVs. 

\subsection{Visual explanations}
\begin{figure}[htp!]
    \centering
    \includegraphics[width=9 cm]{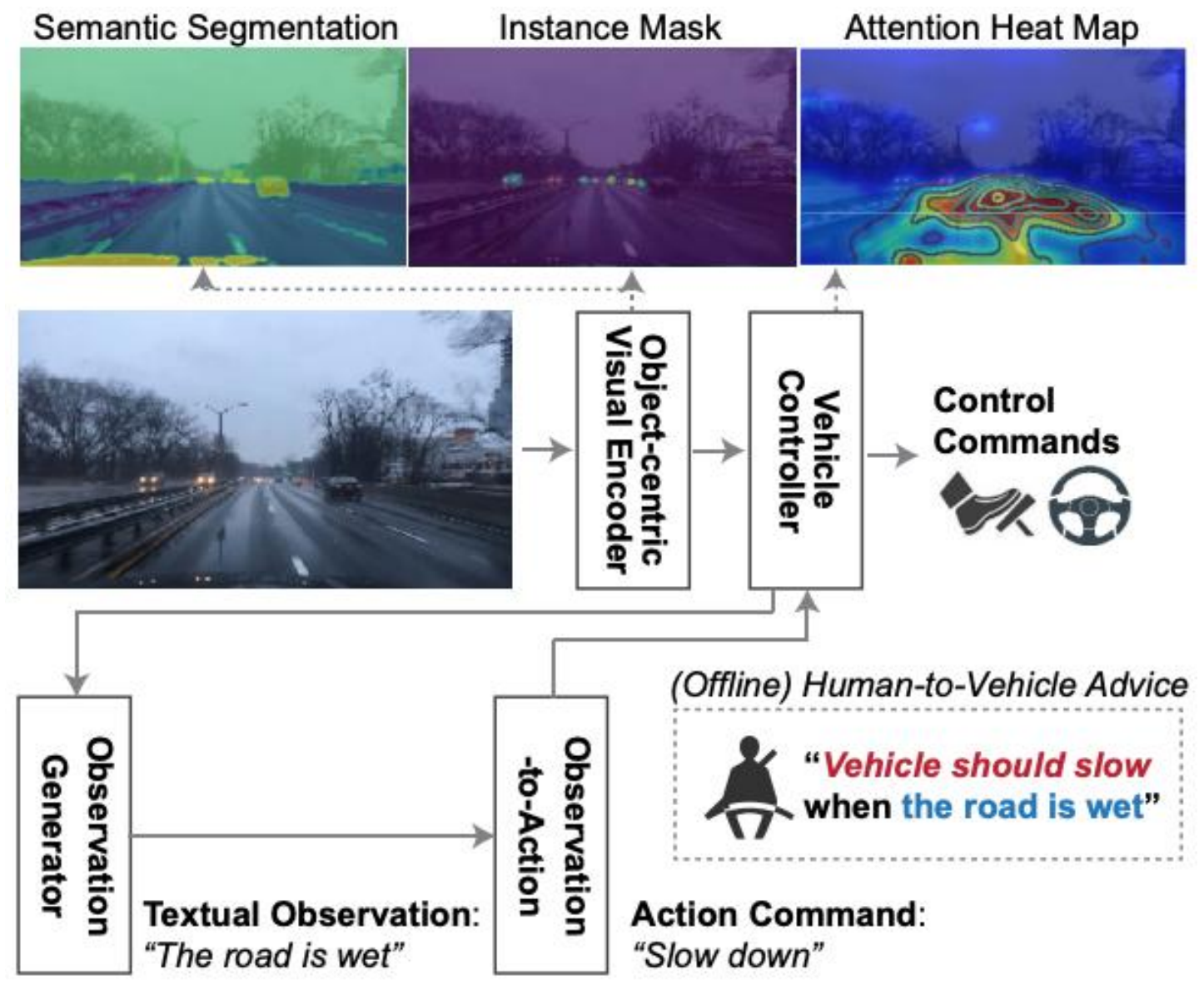}
    \caption{Human advice to the car for relevant action. Source: \cite{kim2020advisable}.}
    \label{fig:action_to_observation}
\end{figure}

As deep neural networks, often in augmented forms of CNNs, power the vision ability of intelligent vehicles, understanding how CNNs capture real-time image segments that lead to the particular behavior of a vehicle is a key concept to achieving visual explanations. In this regard, explainable CNN architectures have resulted in adjustments to generate visual explanations. Zeiler et al. \cite{zeiler2014visualizing} use deconvolution layers to understand the internal representation of CNNs in their seminal work. Hendricks et al. \cite{hendricks2016generating} propose a model concentrating on distinguished properties of objects that explain the rationale for the predicted label. Zhou et al.'s \cite{zhou2016learning} saliency map architecture, class activation map (CAM), highlights the discriminative part of an image to predict the label of the image. Moreover,  Selvaraju et al. \cite{selvaraju2017grad} propose an augmented version of CAM, called Grad-CAM, that highlights the derivative of CNN’s prediction with respect to its input. Further examples of backpropagation-based methods include guided-backpropagation, \cite{springenberg2014striving}, layer-wise relevance propagation \cite{samek2016interpreting, lapuschkin2019unmasking}, and DeepLift \cite{shrikumar2017learning}. Babiker and Goebel \cite{babiker2017introduction, babiker2017using} have also shown that heuristics-based Deep Visual Explanations (DVE) provide a justification for predictions of a CNN. \\
Explaining autonomous driving decisions using visual techniques is also primarily motivated by these studies. Particularly, Bojarski et al.'s work \cite{bojarski2016visualbackprop} is the first explainable vision approach for self-driving, where the authors propose a visualization method, called VisualBackProp, showing which set of \textit{input pixels} contributes to a prediction made by CNNs. Their experiments conducted with the Udacity self-driving car dataset on an end-to-end autonomous driving task show that the proposed technique is a useful tool for debugging predictions of CNNs. \\
Hofmarcher et al. \cite{hofmarcher2019visual} propose a \textit{semantic segmentation model} implemented as a pixel-wise classification
that explains underlying real-time perception of the environment. They evaluate the performance of their framework on Cityscapes \cite{cordts2016cityscapes}, a benchmark dataset for understanding street scenes. The framework outperforms other popular segmentation models such as ENet and SegNet with 59.8 per-class mean intersection over union (IoU) and 84.3 per-category mean IoU. Interpretability of the model is a plus for unexpected behaviors, allowing to debug the driving system and understand the rationales for temporal decisions of a self-driving vehicle. \\
Kim and Canny \cite{kim2017interpretable} use a \textit{causal attention} model on top of the saliency filtering that indicates which input regions actually affect the steering control. Their experiments are conducted on the driving datasets - Comma.ai \cite{CommaAIdataset}, Udacity \cite{Udacitydataset}, and Hyundai Center of Excellence in Integrated Vehicle Safety Systems and Control (HCE): This model runs for nearly 16 hours to train CNNs end-to-end from images to steering angles and apply causality filtering to find out which parts of images have high influence in predictions. With this approach, the learned framework provides an interpretable visualization of a vehicle’s actions. As an enhancement of this model, Kim et al. \cite{kim2018textual} provide textual explanations in their further study. They produce ``intelligible explanations'' on the decisive actions of a self-driving vehicle using an attention-based video-to-text mechanism and introduce a novel dataset, called Berkeley Deep Drive-X (eXplanation) (BDD-X), that contains annotations for textual explanations and descriptions. \\
Zeng et al.'s \cite{zeng2019end} architecture learns to drive an autonomous vehicle safely by following traffic rules, including interaction with road users, yielding, and traffic signals. They use raw LIDAR data and an HD map to generate interpretable representations as 3D detection of objects, anticipated future trajectories, and cost map visualizations. 3D detection instances provide descriptive information so that the model understands the operational environment. Motion forecasting, measured as L1 and L2 distances, explains whether erroneous actions are due to incorrect velocity or
calculation of direction. Finally, Cost Map visualization describes the traffic scene via a top-down view. The architecture is evaluated on a large real-driving dataset consisting of 6,500 traffic scenarios with 1.4 million frames and collected across several cities in North America, and measuring traffic rule violation, closeness to human trajectory, and collision. The authors also carry out an ablation study and show the impact of different overrides, input horizons, and training losses on end-to-end learning. \\
Xu et al. \cite{xu2020explainable} propose \textit{object-induced actions} with explanations for predictions of an autonomous car. The authors introduce a new dataset called BDD-OIA, as an extension of the BDD100K dataset \cite{yu2020bdd100k}; this extension is annotated with 21 explanation templates on a set of 4 actions. Their multi-task formulation for predicting actions also improves the accuracy of action selection. The CNN architecture further unifies reasoning on action-inducing objects and the context of scenes globally. The empirical results of the study on the introduced BDD-OIA dataset show that the explainability of the architecture also enhances action-inducing object recognition, resulting in better self-driving. \\ 
In two respective studies, Kim et al., \cite{kim2020advisable, kim2021toward} propose an approach that leverages \textit{human advice} to learn vehicle control (Figure \ref{fig:action_to_observation}). By sensing operational surroundings, the system is able to generate intelligible explanations on the decisive actions (For example, ``Slowing down \textit{because} the road is wet''). The proposed architecture incorporates semantic segmentation with an attention mechanism that enriches knowledge representation. Experiments performed on the BDD-X dataset show that human advice with semantic segmentation and heat maps improves both the safety and explainability of predictive actions of a self-driving vehicle. \\
As a more recent vision-to-text approach, Atakishiyev et al. \cite{atakishiyev2023explaining} employ the visual question answering (VQA) mechanism to explain autonomous driving actions. They train an RL agent and generate driving data showing the self-driving car’s motion from its field of view. They further convert this video to image sequences, manually annotate the images with question-answer (QA) pairs, and encode questions and images with LSTM \cite{hochreiter1997long} and pre-trained VGG-19 \cite{simonyan2015vgg}, respectively. The experimental results on five action categories show that VQA is a straightforward, effective, and human-interpretable approach to justify autonomous driving actions. Leveraging frontal images for interpretable decision-
making has further been explored by subsequent studies as well \cite{araluce2024leveraging, dong2023did, echterhoff2024driving}. \\
While the mentioned studies focus on vision-based explanations of already obtained predictions of the model, there have been some recent studies paying attention to \textit{counterfactual explanations}. In the context of automated driving, counterfactual analysis can be described with such an exemplary question: “Given the driving scene, how can it be modified so that the vehicle keeps driving instead of stopping ?” In other words, given the input, counterfactual analysis intends to figure out the distinguished features in this input that cause the model to make a certain prediction by envisioning modification of those features would cause the model to make a different prediction (e.g., Figure \ref{fig:counterfactual_explanations}). Thus, in this case, the predictions obtained by the existing model and the imagined model become contrastive. As the application of counterfactual intervention, Li et al. \cite{li2020make} presents an approach to find out risk objects that result in particular driving behavior. Their method, formalized as a Functional Causal Model (FCM), shows that the random elimination of some objects from the scene changes the driving decision to the contrastive prediction, such as from the “Stop” to “Go” command. In further work, Jacob et al. \cite{jacob2022steex} introduce the STEEX model that uses a pre-trained generative model to produce counterfactual rationales by modifying the style of the scene while retaining the structure of the driving scene. Finally, as further enhancement of STEEX,  Zemni et al. \cite{zemni2023octet} propose a method called OCTET that generates object-aware counterfactual
explanations without depending on the structural layout of the driving scene as backpropagation can optimize the spatial positions of the provided instances. \\
Overall, we observe a significant focus on perception-based explanations of autonomous driving systems, as such explanations provide an opportunity to better understand how accurately a self-driving vehicle senses the operational environment. We show a 

\hspace{-0.60cm}
summary of vision-based explanations for AVs in Table \ref{tab:vision_based_explanations}. 

\subsection{Reinforcement learning and Imitation learning-based explanations}
\begin{figure}[htp!]
    \centering
    \includegraphics[width=0.47\textwidth]{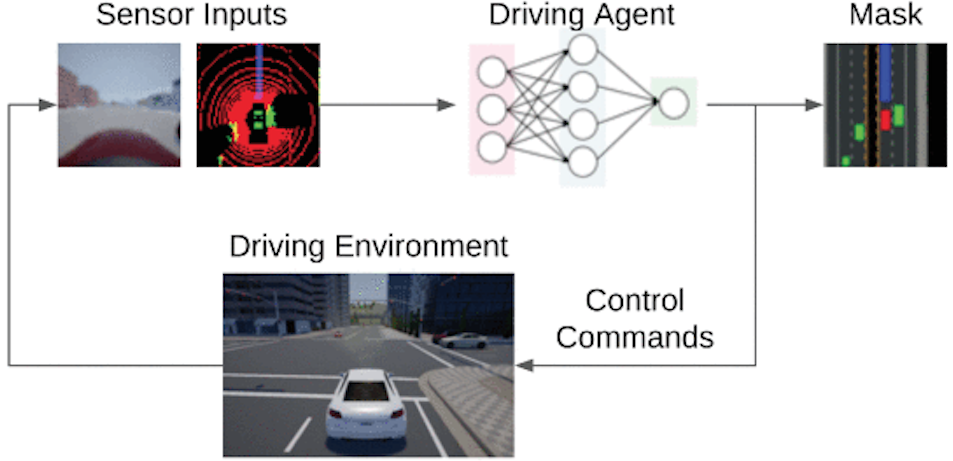}
    \caption{RL-based interpretable end-to-end autonomous driving via a bird-eye mask. Credit: \cite{chen2021interpretable}}
    \label{fig:int_end_to_end_driving}
\end{figure}

Explaining how perceived environmental states are mapped to actions has also recently received attention in the autonomous driving community. In this regard, the field of explainable reinforcement learning (XRL) is a relatively new and emerging research avenue on XAI \cite{heuillet2021explainability, milani2023explainable, bekkemoen2024explainable}. Like vision-based explanations, XRL techniques also aim to provide some forms of justifications on chosen actions of a vehicle either via intrinsically interpretable design or post-hoc explanations. One of the early works in this context is the Semantic Predictive Control (SPC) framework \cite{pan2019semantic}, where the authors propose a data-efficient policy learning approach that predicts future semantic segmentation and provides visual explanations of a learned policy. The framework concatenates multi-scale intermediate features from RGB with tiled actions. The joined modules are then fed into the multi-scale prediction model that predicts future features. Finally, in the last part of the pipeline, the information prediction module inputs the latent feature representation and outputs driving signals alongside the semantic segmentation of the scene.   \\
Chen et al. \cite{chen2021interpretable} introduce a sequential \textit{latent environment model} learned with RL and a probabilistic graphical model-based approach interpreting autonomous cars' actions via a bird-eye mask. They use video cameras and LIDAR images as 
 input in the CARLA simulator \cite{dosovitskiy2017carla}. For the purpose of interpretability of actions and explainability of a learned policy, they generate a bird-eye mask (i.e., Figure \ref{fig:int_end_to_end_driving}). Their model outperforms the used baseline models - DQN, DDPG, TD3, and SAC. Similarly,  Wang et al. \cite{wang2021learning} propose an interpretable end-to-end vision-based motion planning (IVMP) to interpret the underlying actions of a self-driving vehicle. They use semantic maps of a bird-view space in order to plan the motion trajectories of an autonomous car. Moreover, the IVMP approach uses an optical flow distillation network that can improve the real-time performance of the network. The experiments conducted on the nuScenes dataset \cite{caesar2020nuscenes} show the superiority of the proposal over modern approaches in semantic map segmentation and imitation of human drivers.
 In another probabilistic decision-making model, Wang et al. \cite{wang2021uncovering} approach lane merging task as a dynamic process and integrate internal states into joint Hidden Markov Model (HMM) and Gaussian Mixture Regression (GMM). The experiments conducted on the

 \begin{table*}[t!]
\normalsize
\caption{Studies on reinforcement learning and imitation learning-based explanations for AVs}
\captionsetup[table]{position=above}
\centering
\resizebox{\textwidth}{!}{%
    \begin{tabular}{c p{6cm} p{6cm} p{6cm} p{6cm} c}
        \specialrule{.2em}{.1em}{.1em}
        \multirow{3}{*}{\textbf{Study}} & \multicolumn{3}{c|}{\textbf{}} & \multirow{3}{*}{\textbf{Target audience}} \\ 
        & \textbf{Task} & \textbf{Algorithms/Methods} & \textbf{Delivery format} & \\ \specialrule{.2em}{.5em}{.5em}
	Pan et al., \cite{pan2019semantic}, 2019 & Semantic predictive control for explainable policy learning  & LSTM, DDPG-SEG, DLA, model-based RL & Visual & AVs developers  \\
   \hline
			Bansal et al., \cite{bansalchauffeurnet}, 2019 & Leveraging concept
bottlenecks as visual features for predicting control command and explanations of vehicle and human behavior & ChauffeurNet, AgentRNN,
PerceptionRNN, Imitation learning & Visual & AVs developers \\
\hline
Cultrera et al., \cite{cultrera2020explaining}, 2020 & Explaining autonomous driving by learning end-to-end visual attention & CNN, IL & Visual & AVs developers \\

\hline
Chen et al., \cite{chen2021interpretable}, 2021 & Interpretable end-to-end autonomous
driving with latent deep reinforcement
learning & MaxEnt RL, DQN, DDPG, TD3 and SAC & Visual & AVs developers \\

\hline
Schmidt et al., \cite{schmidt2021can}, 2021 & Interpretable and verifiably RL for autonomoous driving
learning & SafeVIPER, PPO & Visual & AVs developers \\

\hline
Wang et al., \cite{wang2021learning}, 2021 & Learning interpretable end-to-end
vision-based motion planning with optical
flow distillation & IVMP, Optical flow & Visual & AVs developers \\

\hline
Wang et al., \cite{wang2021uncovering}, 2021 & Uncovering interpretable internal states of
merging tasks at highway on-ramps for
autonomous driving decision-making & GMR, HMM & Visual & AVs developers \\

\hline
Rjoub et al., \cite{rjoub2022explainable}, 2022 & XAI-based federated deep RL for
autonomous driving & DQN, DQN-XAI, SHAP & Visual & AVs developers \\

\hline
Renz et al., \cite{renz2022plant}, 2022 & Explainable planning for autonomous
driving & BERT, GRU, Imitation Learning & Visual & AVs developers \\

\hline
Teng et al., \cite{renz2022plant}, 2022 & Interpretable imitation learning for
end-to-end autonomous driving & Bird’s Eye View model, Imitation Learning & Visual & AVs developers \\

\hline
Hejase et al., \cite{hejase2022dynamic}, 2022 & Interpretable state representation for
deep RL in autonomous driving & DDQN & Visual & AVs developers \\

\hline
Cultrera et al., \cite{cultrera2023explaining}, 2023 & Visual attention and end-to-end trainable region proposals for explainable autonomous driving & IL, Region Proposal Networks, Spatial Transformers Network & Visual & AVs developers \\

\hline
Shao et al., \cite{shao2023safety}, 2023 &  Interpretable sensor fusion transformer for safe autonomous driving & InterFuser, Imitation Learning, ResNet, GRU & Visual & AVs developers \\

\hline
Paleja et al., \cite{palejalearning}, 2023 & Interpretable continuous
control trees for autonomous driving & Differentiable Decision Trees, SAC & Visual & AVs developers \\

\hline
Kenny et al., \cite{kenny2023towards}, 2023 & Interpretable Deep RL with Human-Friendly Prototypes for autonomous driving & PW-Net, PPO, TD3 & Visual & AVs developers \\

\hline
Yang et al., \cite{yang2023leveraging}, 2023 & Reward consistency for interpretable
feature discovery for autonomous driving & PPO & Visual & AVs developers \\

\hline
Lu et al., \cite{lu2024enhancing}, 2024 & Human-like cognitive maps for enhancing interpretability of autonomous driving  & Successor Representations, Cognitive Potential Field & Visual & AVs developers \\

\hline
Liu et al., \cite{liu2024interpretable}, 2024 & Interpretable generative adversarial IL for autonomous driving  & IL, Signal Temporal Logic & Visual & AVs developers \\

			\specialrule{.2em}{.1em}{.1em}
   \label{tab:rl_il_exp}
		\end{tabular}%
	}
\end{table*} \hspace{-0.62cm}
INTERACTION dataset \cite{zhan2019interaction} demonstrate the efficiency of the proposed technique and show that merging at highway on-ramps can be delineated by three interpretable internal states in terms of the absolute speed of a vehicle while merging. \\
Rjoub et al. \cite{rjoub2022explainable} have shown that federated deep RL combined with XAI can lead to trusted autonomous driving. They use a federated learning approach for decision-making and leverage edge computing that enables different devices to train an ML model in a collaborative manner. The model is first developed on the parameter server and further broadcasted to other devices. Then, global ML methods intake updates from these devices regularly, and the process continues until the model performs well enough on driving tasks. Yang et al. \cite{yang2023leveraging} have also shown that an intrinsically interpretable RL agent for autonomous driving can be achieved via reward consistency aiming to resolve the gradient disconnection in reward-action mapping. \\
Finally, within the exploration via interaction context, a few studies have employed various forms of imitation learning (IL) techniques for explainable autonomous driving. Cultrera et al. \cite{cultrera2020explaining} present conditional imitation learning with an end-to-end visual attention model, which identifies those parts of images that have a higher influence on predictions. They test their architecture in the CARLA simulator on four tasks - go straight, turn left, turn right, and follow the lane. Their ablation study focused on box type importance and fixed grid analysis to get an attention map on the images shows that integrated imitation learning and attention model enables a car to drive safely and perform relevant maneuvers in real-time. \\
Leveraging vision for interpretability of an agent's actions, Teng et al. \cite{teng2022hierarchical} propose hierarchical interpretable IL (HIIL) technique

\hspace{-0.62cm}
that unifies bird eye view (BEV) mask with the steering angle to perform actions in complex situations as an end-to-end autonomous driving pipeline. They construct their method as a two-phase task: In the first phase, the pre-trained BEV model is used to interpret the driving environment. Then imitation learning takes the latent features of BEV mask from the first phase and combines them with a steering angle acquired through the Pure-Pursuit algorithm. The experiment performed in the CARLA simulator shows that the proposed method enhances the interpretability and robustness of driving in various circumstances. \\
Moreover, Renz et al. \cite{renz2022plant} introduce PlanT, a rigorous IL-based approach that uses transformers for planning. PlanT is able to explain its action decision by recognizing the most important object in its driving segment and outperforms state-of-the-art work in CARLA's Longest6 Benchmark by 10 points (See \cite{PlanT2022} for a

\begin{table*}[t!]
\normalsize
\caption{Studies on decision tree-based explanations for AVs}
\captionsetup[table]{position=above}
\centering
\resizebox{\textwidth}{!}{%
    \begin{tabular}{c p{6cm} p{6cm} p{6cm} p{6cm} c}
        \specialrule{.2em}{.1em}{.1em}
        \multirow{3}{*}{\textbf{Study}} & \multicolumn{3}{c|}{\textbf{}} & \multirow{3}{*}{\textbf{Target audience}} \\ 
        & \textbf{Task} & \textbf{Algorithms/Methods} & \textbf{Delivery format} & \\ \specialrule{.2em}{.5em}{.5em}
   Omeiza et al., \cite{omeiza2021towards}, 2021 & Generating tree-based explanations with
and without causal attributions  & Tree-based representation / User study & Textual & All groups  \\
   \hline
         Brewitt et al., \cite{brewitt2021grit}, 2021 & Interpretable and verifiable goal
recognition with learned decision trees for
autonomous driving & Decision Tree & Visual and Textual & AVs developers \\

\hline
         Mankondiya et al., \cite{mankodiya2021xai}, 2021 & XAI for trust management in autonomous
vehicles & Random Forest, Decision Tree, AdaBoost & Visual & AVs developers \\
\hline
   Cui et al., \cite{cui2022interpretation}, 2022 & Interpretation framework for autonomous
driving & Random Forest, SHAP & Visual & AVs developers \\

\hline
   Brewitt et al., \cite{brewitt2023verifiable}, 2023 & Interpretatable trees for goal recognition in autonomous driving under occlusion & OGRIT, Decision Tree & Visual & AVs developers \\

         \specialrule{.2em}{.1em}{.1em}
   \label{tab:tree_exp}
      \end{tabular}%
   }
\end{table*} \hspace{-0.62cm}
visual demonstration). In a more recent work in this context, Liu et al. \cite{liu2024interpretable} show that combining Signal Temporal Logic with generative IL is also an effective approach for interpretable policy for autonomous driving. Overall, while building interpretable by-design RL or IL agent for autonomous driving is a challenging task, employing external methods, such as logic and vision, can help achieve explainability for the agent's actions. A summary of RL and IL-based explanations is shown in Table \ref{tab:rl_il_exp}. 

\subsection{Decision tree-based explanations}
Being inherently interpretable and easier to understand a prediction of a model, decision-tree-based explanations have also been investigated in autonomous driving. Decision trees have been proven to describe the rationale semantically for each prediction made by a CNN architecture \cite{Zhang_2019_CVPR}. Omeiza et al. \cite{omeiza2021towards} use decision trees as a \textit{tree-based} representation that generates scenario-based explanations of different types by mapping observations to actions in accordance with traffic rules. They use human evaluation in a variety of driving scenarios and generate Why, Why Not, What If, and What explanations for driving situations and empirically prove that the approach is effective for the intelligibility and accountability goals of automated vehicles. \\
Brewitt et al. \cite{brewitt2021grit} introduce  Goal Recognition with Interpretable
Trees (GRIT), a framework that uses decision trees trained from the trajectory data of a self-driving car. The framework, tested on fixed-frame scenarios, is proven empirically verifiable for goal recognition using a satisfiability modulo theories (SMT) solver \cite{de2008z3}.\\
Cui et al. \cite{cui2022interpretation} use Random Forest for the interpretability purpose on the autonomous car-following task. They employ deep RL for the decision-making of an autonomous car and employ SHAP values to simplify the feature space. Once the agent generates state-action pairs, Random Forest is applied to these pairs and experimental results show the approach works effectively to explain behavior for the designated car-following task. In a recent study, Random Forest has also been proven to detect misbehaving vehicles in Vehicular Adhoc Networks (VANET) in Mankodiya et al.'s work \cite{mankodiya2021xai}. Thus, being computationally more transparent than traditional deep neural network architectures, decision trees can explain behaviors of a variety of autonomous driving tasks with less computation.

\subsection{Logic-based explanations}
While the interpretability of a deployed autonomous driving control model has been the dominant direction for research, there have also been attempts to verify the safety of self-driving vehicles with logical reasoning. In this regard, Corso and Kochenderfer \cite{corso2020interpretable} present a technique to identify interpretable failures of autonomous cars. They use \textit{signal temporal logic} expressions to
describe failure cases of an autonomous car in unprotected left turn and pedestrian crossing scenarios. For this purpose, the authors use genetic programming to optimize signal temporal logic expressions that acquire disturbances trajectories causing a vehicle to fail in its decisive action. The experimental results show that the proposed approach is effective in interpreting the safety validation of a car. \\
Suchan et al.  \cite{suchan2019out} have developed an \textit{answer set programming}-based abductive reasoning framework for online sensemaking for perception and control tasks. In its essence, the framework integrates knowledge representation and computer vision in an online manner to explain the dynamics of traffic scenes, particularly occlusion scenarios. The authors demonstrate their method's explainability and commonsensical value with empirical study collected on the KITTI MOD \cite{geiger2012we} dataset and the MOT benchmark \cite{milan2016mot16}. Another experimental study leveraging the concept of answer set programming has been carried out by  Kothawade et al. \cite{kothawade2021auto}: they introduce AUTO-DISCERN, a system that incorporates common sense reasoning with answer set programming to automate explainable decision-making for self-driving vehicles. They test their rules and show AUTO-DISCERN's credibility in real-world scenarios, such as lane changing and right turn operations, from the KITTI dataset.\\

\subsection{User study-based explanations}
Some investigations involve users in case studies to understand the effective strategies for explanation generation in autonomous driving tasks. The key idea of a user study is that getting people's input in designated driving tasks can help improve the adequacy and quality of explanations in autonomous driving. Wiegand et al. \cite{wiegand2019drive} perform a user study that identifies a mental model of users for determining an effective practical implementation of an explanation interface. The main research question here is to understand what components need to be visualized in a vehicle so the user can comprehend the decisions of self-driving vehicles. The study discloses that combining an expert mental model with a user mental model as a target mental model enhances the situation awareness of the drivers. Furthermore, Wiegand et al. \cite{w2020d} investigate situations, in which explanations are needed and methods pertinent to these situations. They spot seventeen scenarios where a self-driving vehicle behaves unexpectedly. Twenty-six participants are selected to validate these situations in the CarMaker driving simulator to provide insights into drivers’ need for explanations. As a result of the user study, the authors identify six groups to highlight the primary concerns of drivers

\begin{table*}[t!]
\normalsize
\caption{Studies on logic-based explanations for AVs}
\captionsetup[table]{position=above}
\centering
\resizebox{\textwidth}{!}{%
    \begin{tabular}{c p{6cm} p{6cm} p{6cm} p{6cm} c}
        \specialrule{.2em}{.1em}{.1em}
        \multirow{3}{*}{\textbf{Study}} & \multicolumn{3}{c|}{\textbf{}} & \multirow{3}{*}{\textbf{Target audience}} \\ 
        & \textbf{Task} & \textbf{Algorithms/Methods} & \textbf{Delivery format} & \\ \specialrule{.2em}{.5em}{.5em}
 Suchan et al., \cite{suchan2019out}, 2019 & An answer set programming-based abductive reasoning for visual sensemaking & Answer set programming, YOLOv3, SSD, Faster R-CNN & Visual & AVs developers \\
      \hline
	Corso and Kochenderfer \cite{corso2020interpretable}, 2020 & Interpretable safety validation for autonomous driving  & Signal temporal logic & Textual & AVs developers  \\

\hline
	DeCastro et al., \cite{decastro2020interpretable}, 2020 & Interpreting policies via signal temporal logic for autonomous driving & Signal temporal logic, LSTM, CVAE & Visual & AVs developers \\
 
 \hline
			Kothawade et al, \cite{kothawade2021auto}, 2021 & Explainable autonomous driving using commonsense reasoning  & ASP, s(CASP) & Textual & Road users \\

			\specialrule{.2em}{.1em}{.1em}
		\end{tabular}%
	}
\end{table*}\hspace{-0.62cm}
with these unexpected behaviors, namely emotion and evaluation, interpretation and reason, the capability of a self-driving car, interaction, driving forecasting, and request times for explanations. \\
Wang et al. \cite{wang2021human} propose an approach that enables a human driver to provide \textit{scene forecasting} to an intelligent driving system using a purposeful gaze. They develop a graphical user interface to understand the effect of human drivers on the prediction and control of an intelligent vehicle. A simulator is used to test and verify three driving situations where a human driver’s input can improve safety of self-driving.\\
Apart from these works, Schneider et al. involve human participants in their empirical studies to understand the role of explanations for the public acceptance of AVs \cite{schneider2021increasing, schneider2021explain}. They explore the role of explainability-supplied UX in AVs, provide driving-related explanations to end users with different methods, such as textual, visual, and lighting techniques, and conclude that providing context-aware explanations on autonomous driving actions increases users’ trust in this technology. Their subsequent study also confirms that driving explanations can help mitigate the negative impact of AVs failures on users \cite{schneider2023don}. Finally, Kim et al.'s user study \cite{kim2023and} confirms that humans do not need explanations seamlessly, and presenting explanations only in critical driving conditions is preferred to enjoy the trip with an autonomous car and prevent information overload. We summarize user studies on XAI for AVs in Table \ref{tab: user_study_exp}.

\subsection{Large Language Models and Vision-Language Models-based explanations}
Finally, while the preliminary studies and further works on explainable autonomous driving primarily focused on a combination of various AI techniques revisited above, large language models (LLMs) and vision-language models (VLMs) have recently emerged as a novel paradigm to interpreting AVs decisions and describing traffic scenes. Built on top of Foundation Models, such as  GPT \cite{radford2018improving} and BERT \cite{devlin2019bert}, there has been significant progress on building both domain-agnostic and domain-specific LLMs (e.g., GPT-3 \cite{brown2020language}, GPT-4 \cite{achiam2023gpt}, LLAMA \cite{touvron2023llama}, LLAMA-2 \cite{touvron2023llama2}, Vicuna \cite{vicuna2023}, Alpaca \cite{taori2023stanford}, Claude \cite{anthropic_claude}) and VLMs (e.g., Flamingo \cite{alayrac2022flamingo}, LLaVa \cite{liu2024visual}, PaLM-E \cite{driess2023palm}, Video-LLaVA \cite{lin2023video}, Video-LLaMA \cite{zhang-etal-2023-video}, Gemini \cite{team2023gemini}, Claude 3 \cite{anthropic_claude3}). In this sense, there have been tremendous efforts to build language and vision-language models on top of these base models for interpretable autonomous driving. Overall, based on the recent trend in leveraging these large models for the interpretability purpose in autonomous driving, we observe the following directions: \\
\textit{Presenting live natural language explanations during the trip}: The  promising work in this context is Wayve’s LINGO-1 \cite{wayve_lingo1_2023} and LINGO-2 \cite{wayve_lingo2_2024} architectures. LINGO-1 is a vision-language-action (VLAM) model that provides live natural language explanations for describing a vehicle's chosen actions in end-to-end autonomous driving. Trained on diverse multimodal (vision and language) datasets, LINGO-1 can describe action decisions and causal factors inducing these actions. The advantage of LINGO-1 architecture is that its explanations are concise, informative, and reflect temporal changes in the driving environment. The Wayve team has also achieved live linguistic explanations for autonomous driving in a simulation environment \cite{chen2023driving}.\\
\textit{Video Question Answering as a reasoning technique}: An essential characteristic of modern AVs is the consideration of human factors in the design and development of this technology and having effective human-machine alignment for trustworthy autonomous driving. In this sense, it is crucial that users-in-the-loop have some form of interaction with AVs during a journey. Motivated by this concept, some recent works have approached conversational user interface between people on board and AVs as a Video Question Answering (VideoQA) task \cite {xu2023drivegpt4, marcu2023lingoqa, park2024vlaad}. Asking questions about the behavior of an autonomous system is a part of our intuition, and in the context of AVs, getting answers to traffic-related situations and autonomous car action-related questions can help users have a comfortable and reliable journey. Other practical applications of LLMs and VLMs are interpretable motion planning \cite{mao2023gpt}, chain-of-thought-based reasoning for control and decision-making \cite{sha2023languagempc, wen2023dilu}, action justification with a control signal \cite{yuan2024rag}, predicting the intent of other traffic actors via visual reasoning and cues \cite{dewangan2023talk2bev}, and understanding the role of video transformer based-explanations on safety of autonomous driving \cite{atakishiyev2024safety} (see Table \ref{tab:llm_vlm_exp} for more details on LLM and VLM-based explanations). \\
Overall, as an emerging AI technology, LLMs and VLMs have tremendously benefitted AVs from the interpretability aspect, as described in the above studies. However, it is also worthwhile to mention that there are still spaces for improvement of these models as fictitiously generated explanations (e.g., \textit{hallucinations}) may have serious safety implications and high-stakes consequences for self-driving actions and human life. We describe such caveats and potential solutions in Section 6 as future work. \\  
A high-level overview of all these studies indicates driving explanations are generally multi-modal, context-dependent, and task-specific. Moreover, end-to-end learning has become even more popular for highly autonomous decision-making owing to the combination of powerful deep-learning approaches and overall safety benefits. Based on the insights from the reviewed literature, we can define explainable autonomous driving as follows: 
\begin{quote}\label{XAI_def} 
\textit{Explainable autonomous driving is a self-driving approach powered by a compendium of AI techniques 1) ensuring an acceptable level of safety for a vehicle's real-time decisions, 2) providing explanatory information on the action decisions in critical traffic scenarios in a timely manner, and 3) obeying all traffic rules established by the legal entities and regulators.} 
\end{quote} 
\begin{table*}[t!]
\normalsize
\caption{User study-based explanations for AVs}
\captionsetup[table]{position=above}
\centering
\resizebox{\textwidth}{!}{%
    \begin{tabular}{c p{6cm} p{6cm} p{6cm} p{6cm} c}
        \specialrule{.2em}{.1em}{.1em}
        \multirow{3}{*}{\textbf{Study}} & \multicolumn{3}{c|}{\textbf{}} & \multirow{3}{*}{\textbf{Target audience}} \\ 
        & \textbf{Task} & \textbf{Algorithms/Methods} & \textbf{Delivery format} & \\ \specialrule{.2em}{.5em}{.5em}
 Wiegand et al., \cite{wiegand2019drive}, 2019 & Explaining driving behavior of
autonomous cars  &  User study & Textual & Backup drivers  \\
   \hline
         Wiegand et al., \cite{w2020d}, 2020 & Understanding situations that a driver needs explanations & User study & Visual & All groups \\

\hline
         Wang et al., \cite{wang2021human}, 2021 & Enhancing automated driving with human foresight & User study & Visual & Backup drivers \\
\hline
   Schneider et al., \cite{schneider2021explain}, 2021 & UX for transparency in autonomous driving & UEQ-S, AVAM (User study) & Visual, Textual, Light & All groups \\

\hline
   Schneider et al., \cite{schneider2021increasing}, 2021 & Increasing UX through different feedback modalities & UEQ-S (User study) & Visual, Textual, Audio, Light, Vibration & All groups \\

 \hline
   Shen et al., \cite{shen2022to}, 2022 & Identifying which scenarios need explanations in autonomous driving & Friedman test, Pearson correlation, Point-Biserial Correlation & Visual & Road users \\

 \hline
   Schneider et al., \cite{schneider2023don}, 2023 & The role explanatory information in failure situations in highly autonomous driving & UEQ-S, AVAM & Visual, Textual & All groups \\
 \hline
   Kim et al., \cite{kim2023and}, 2023 & Timing perspective and mode of explanations for road users in autonomous driving & GradCam, Head-mounted display, Windshield display & Visual & Road users \\

         \specialrule{.2em}{.1em}{.1em}
   \label{tab: user_study_exp}
      \end{tabular}%
   }
\end{table*}
Driven by this definition and the state-of-the-art works in the above sections, we present a general and conceptual XAI framework for end-to-end autonomous driving aligned with industrial trends and show the necessary components and process steps to achieve regulatory-compliant AVs in the next generation.

\section{An XAI framework: integrating end-to-end control, safety, and explanations}
We present a general framework in which methods for developing XAI, end-to-end learning, and safety components are combined to inform processes of regulatory principles. Each of these components has a concrete role in our framework. In our recent study \cite{atakishiyev2023towards}, we have covered a brief description of end-to-end learning for AVs. We extend the scope of that work and describe the essential elements of end-to-end autonomous driving, and the role of and potential challenges with explanations in such a setting. We describe these individual components as follows:\\\\
\textbf{1. An end-to-end control component}: Given all  possible instances of environment, \[E = \{e_1, e_2, ... e_n\},\] and a compendium of actions \[A = \{a_1, a_2, ... a_n\},\]
an autonomous car can take, the overall role of a\textit{ control system} is to map the perceived environment to corresponding actions:
                            			
\[C: E \mapsto  A.\]

This mapping intends to ensure that a controller maps the environment to a relevant action of an autonomous system. A control system $C$ is an {\it end-to-end control system} ({\it\bfseries eeC}),  if $C$ is a total function that maps every instance of an environment 
\[ e \in E\] 
to a relevant action  \[a \in A.\]
The two most prevalent learning paradigms for end-to-end autonomous driving are RL and IL \cite{chen2023end}. Typically, RL approaches map sensor information to states and produce control signals for an autonomous car. On the other hand, in IL-based end-to-end learning, an agent is trained by imitating an expert's behavior to learn the optimal policy. In practice, both types of IL - behavior cloning and inverse RL - have been applied to various end-to-end driving tasks as described in the above-mentioned studies.\\
\textbf{2. A safety-regulatory compliance component:}
The role of the safety-regulatory compliance component, {\it\bfseries srC}, is to represent the function of a regulatory agency, one of whose main roles is to verify the safety of any combination of $eeC$ with autonomous vehicle actions $A$:
\[srC = f(eeC, A).\]
This requirement could be as pragmatic as some inspection of individual vehicle safety (for example, verifying basic safety functions of an individual vehicle for re-licensing). That said, this concept should be considered as a thorough compliance testing of $eeC$ components from vehicle manufacturers to certify their public safety under international and/or national transportation guidelines such as \cite{regulation2016} and \cite{guideleinescanada}.
The general principles for acceptable functional safety of road vehicles are defined by the ISO 26262 standard \cite{ISO26262}. According to this standard, there should be a safety certification development with evidence-based rationales: the vehicle should be able to meet the established functional safety requirement in its operational context. Part 6 of the ISO 26262 standard \cite{ISO26262_6} is dedicated to end-product development for automotive applications within the software level. This guideline includes the design, development, testing, and verification of software systems in automotive applications. 
Based on these standards, there seem to be two fundamental approaches to confirming regulatory compliance, which we label confirmation of compliance by ``simulation,'' and confirmation of compliance by ``verification,''. These steps are aligned with our observation regarding the role of XAI in confirming regulatory compliance. In the case of the process of establishing regulatory compliance by simulation, the idea is that a selected set of autonomous actions can be simulated, and then assessed to be satisfactory. This approach is perhaps the most familiar, as it arises naturally from an engineering development trajectory, where the accuracy of simulators determines the quality of compliance (e.g., \cite{johansen2016ship}). The confidence of the established compliance is a function of the accuracy and coverage of the simulation. However, this compliance process can be very expensive and prone to safety gaps, especially when consensus on the properties and scope of a simulation is difficult to achieve. Thus, in general, the simulation part can be considered a "driving school" for AVs: The designed and developed learning software system should be tested rigorously

\begin{table*}[t!]
\normalsize
\caption{Studies on large language models and vision-language models-based explanations for AVs}
\captionsetup[table]{position=above}
\centering
\resizebox{\textwidth}{!}{%
    \begin{tabular}{c p{6cm} p{6cm} p{6cm} p{6cm} c}
        \specialrule{.2em}{.1em}{.1em}
        \multirow{3}{*}{\textbf{Study}} & \multicolumn{3}{c|}{\textbf{}} & \multirow{3}{*}{\textbf{Target audience}} \\ 
        & \textbf{Task} & \textbf{Algorithms/Methods} & \textbf{Delivery format} & \\ \specialrule{.2em}{.5em}{.5em}
    Dewangan et al., \cite{dewangan2023talk2bev}, 2023 & Language-augmented Bird’s-eye View Maps for
Autonomous Driving&  GPT-4, GRIT & Visual, Textual & All groups  \\
   \hline
            Xu et al., \cite{xu2023drivegpt4}, 2023 & VQA and natural language-based explanations for autonomous driving & LLAMA 2, CLIP & Visual, Textual & All groups \\

\hline
            Marcu et al., \cite{marcu2023lingoqa}, 2023 & Video question answering for autonomous driving & Vicuna-1.5-7B, GPT-4 & Textual & All groups \\
\hline
    Chen et al., \cite{chen2023driving}, 2023 & Improving context understanding in autonomous driving with object-level vector modalities and LLM & GPT 3.5, PPO & Visual, Textual & All groups \\

\hline
    Fu et al., \cite{fu2023drive}, 2023 & Understanding traffic situations in a closed loop & GPT 3.5 & Textual & All groups \\

 \hline
    Mao et al., \cite{mao2023gpt}, 2023 & Interpretable motion planning as language modeling & GPT 3.5 & Textual & Road users \\

 \hline
    Sha et al., \cite{sha2023languagempc}, 2023 & LLM as a decision-maker in complex driving scenarios & ChatGPT, MPC & Visual, Textual & AVs developers\\

\hline
 Wayve Team \cite{wayve_lingo1_2023}, 2023 & Providing live explanations in natural language & Integrated vision, language and action architecture & Textual & All groups\\

 \hline
 Nie et al., \cite{nie2023reason2drive}, 2023 & Interpretable reasoning in complex driving situations in autonomous driving & GPT-4, MLP, ViT-G/14 & Textual & All groups\\

 \hline
 Park  et al., \cite{park2024vlaad}, 2024 & Video question answering for traffic scene understanding & Video-LLAMA, GPT-4 & Textual & All groups\\
\hline
 Wen et al., \cite{wen2023dilu}, 2024 & LLM-based knowledge-driven approach for interpretable autonomous driving & Out-of-box LLM & Visual, Textual & AVs developers \\

 \hline
 Yuan  et al., \cite{yuan2024rag}, 2024 & Retrieval-augmented VLM for explainable autonomous driving & LanguageBind, MLP, Vicuna-1.5-7B & Textual & All groups\\

 \hline
 Chi  et al., \cite{chi2024multi}, 2024 & GPT-aided explainable decisions for autonomous vehicles & GPT, Graph of Thoughts & Textual & All groups\\

 \hline
 Atakishiyev et al., \cite{atakishiyev2024safety}, 2024 & Robustness of a transformer-based VideoQA model against human-adversarial questions and its safety implications for self-driving  & Video-LLaVA & Textual & All groups\\

 \hline
 Duan et al., \cite{duan2024prompting}, 2024 & Unifying imitation learning with LLMs to enhance safety of end-to-end driving  & Vicuna LLM, Swin transformer & Textual & All groups\\

            \specialrule{.2em}{.1em}{.1em}
            \label{tab:llm_vlm_exp}
        \end{tabular}%
    }
\end{table*}\hspace{-0.62cm}
in this phase before such an autonomous system, as a holistic architecture, is deployed to a real vehicle in the physical environment and real roads. \\ 
The alternative, verification, is aligned with our own framework and has significant foundational components established in the discipline of proving software correctness, with a long history (e.g., \cite{softwarecorrectness1976}). The general idea is that offline simulation-based autonomous driving is validated on real roads on real AVs via real sensor suites and a learning software stack by passing the safety checks of regulatory compliance. \\
In addition to safety assurance, another critical requirement of AVs is their ability to defend against adversarial attacks. The ISO/SAE 21434 standard has defined guidelines for cybersecurity risk management for road vehicles, and AVs must also comply with these requirements \cite{ISO_cybersecurity}. As AVs increasingly rely on their automation ability, it is vital that ML software of an intelligent driving system and built-in interfaces can detect and defend against potential cyber-attacks of the broad spectrum, such as electronic control units (ECU) attacks, in-vehicle network attacks, and automotive key-related attacks   \cite{qayyum2020securing,kim2021cybersecurity, sun2021survey}. \\    
We can expect that the potential evolution of the $srC$ processes will ultimately rely on the automation of regulatory compliance testing against all $eeC$ systems. The complexity of $srC$ systems lies within the scope of the testing methods established in a legal
framework, where these methods are the basis for confirming a threshold of safety. For instance, a regulatory agency may require at least 90\% regulatory-compliant performance of any particular $eeC$ from $N$ safety tests to be performed. However, as a general requirement, this performance must meet ISO 26262 and ISO/SAE 21434 standards to ensure that an autonomous car's decision-making procedure is aligned with its underlying ML software: The safety features must pass critical checkpoints, and the autonomous car has to have the ability to defend itself against foreseeable adversarial attacks.\\
\textbf{3. An explanation component:} This constituent of the framework provides understandable insights into the real-time action decisions made by autonomous driving, complying with and corresponding to an $eeC$ and a $srC$. The explanation component must justify how the autonomous car chooses actions along the trip and has to be able to communicate these pieces of information to the relevant users both during the journey and via a post-trip analysis.  As analyzed in the reviewed studies, explanations can be described in visual, textual, feature importance format or in hybrid, multi-modal ways and conveyed via light, audio, vibrotactile, and in other forms as needed. \\
\textit{Temporal granularity and conveyance of explanations:} While the format and content of explanations have been the primary focus of XAI research, it is noteworthy to underscore that another important consideration, the time granularity of explanations, has not been well-studied in the state of the art. In general, the timing perspective of AVs explanations can be analyzed in three ways: 1) Should explanations be delivered before action is chosen or after action is performed? 2)  What is the appropriate lead time for a safe transition from an automatic mode to a human takeover? and 3) Should explanations be delivered seamlessly or only when it is required? We analyze these nuances separately as follows: \\
1) \textit{Timing mechanism of explanations}: Delivering timely explanations can help human drivers/passengers react to emergent situations, such as takeover requests, appropriately and prevent a potential danger in the vicinity. According to Koo et al.’s study \cite{koo2015did}, it is favorable to convey explanations before a driving event is about to happen. This concept has further been validated by Haspiel et al.’s user study, and human judgment shows that explanations should be delivered \textit{before} action is \textit{decided} rather than \textit{after} it is \textit{performed} \cite{haspiel2018explanations}. This judgment makes sense as on-time communication of explanations can bring situation awareness for people on board and enable them to monitor an autonomous car’s subsequent action. If the action to be performed soon is hazardous, a human driver or passenger can manually intervene in the situation with such explanations and prevent a potential danger ahead. \\
\begin{figure*}[htp!]
    \centering
    \includegraphics[width=0.95\textwidth]{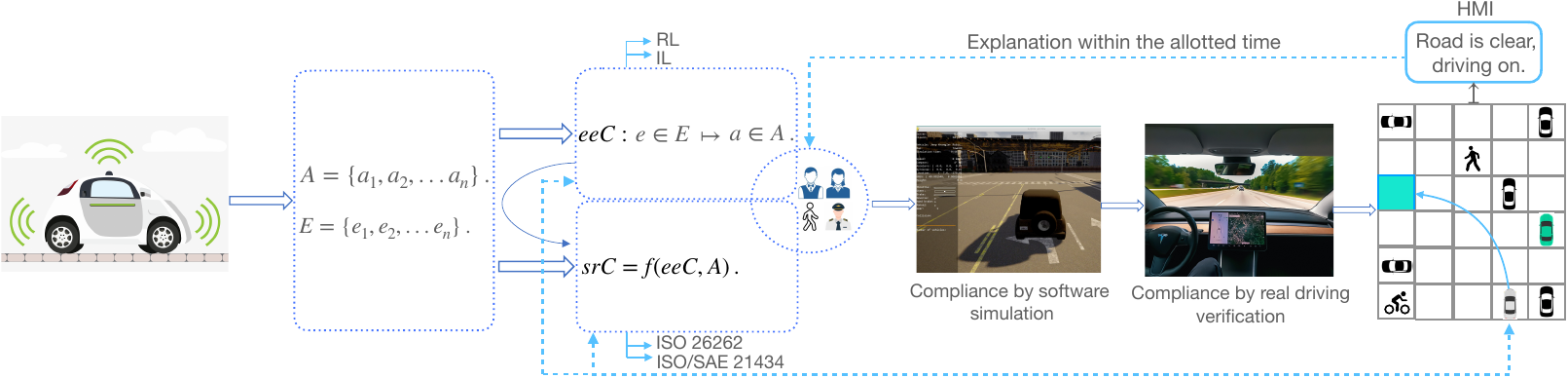}
    \caption{A diagram of the proposed explainable end-to-end autonomous driving framework.}
    \label{fig:end-to-end_control_framework}
\end{figure*}
2) \textit{The impact of lead time on the safe transition from an automated mode to a human takeover}: Another important criterion is determining the amount of time needed to alert human actors for a takeover request. In the user study measuring the impact of 4 s vs. 7 s as the lead time on takeover alert,  Huang and Pitts \cite{huang2022takeover} show that a shorter lead time leads to a faster transition to human-controlled mode but also lacks the quality of takeover as lack of time may be stressful for a human actor in such situations. A similar conclusion has been acquired by Mok et al. \cite{mok2015emergency} in the case of 2, 5, and 8 s transition times. Wan and Xu \cite{wan2018effects} have further verified that an insufficient amount of lead time, such as 3 s, results in an impaired takeover performance, and drivers perform better when enough time, such as more than 10 s, is allotted for takeover requests. In general, it can be concluded that lead time for explaining emergent situations to a human and transitioning control should happen within a few seconds, while for non-critical situations, such as post-trip analysis, the amount of time may be as long as needed. \\
3) \textit{All-time or only necessary explanations?}: It is also important to consider that humans need to enjoy their trips with AVs and get information from a vehicle when it is necessary. This aspect also applies to the delivery of explanatory information to end users. When the passengers/human drivers are provided with tons of information during the trip, it can lead to mental overload for them \cite{w2020d}. Consequently, it is generally favored that driving decisions and traffic scenes may be described to humans on board when traffic conditions are critical, and people need to be alerted.\\
It is also noteworthy to specify that AVs must be equipped with need-based HMIs to deliver explanations. There are some challenges with effective automotive HMI design. First, people may have different choices or preferences for HMI (i.e., display monitor, alert interface, etc.). Furthermore, users' various cognitive and functional abilities must be a crucial factor in the design of user interfaces \cite{arfini2023design}. For instance, people with visual or hearing impairments may need a customized HMI. Hence, automotive manufacturers must consider the diversity of users, contemplate the timing perspective of HMI explanations in line with relevant actions, and reach a consensus on the best practices with effective HMI design for AVs \cite{schieben2019designing}. \\
Based on the mentioned process steps and crucial elements, we see that achieving the interpretability of self-driving models is challenging, necessitating integration of those steps and cooperation between users and AVs. Consequently, while we argue that transparent and highly autonomous driving is feasible, human factors must be a vital consideration in the design and development of such systems. A simple graphical illustration of our proposed framework with its elements can be seen in Figure \ref{fig:end-to-end_control_framework}.\\
In the next section, we envision the future of AVs research in the realm of safety and explainability based on current industrial trends and list some potential challenges toward this goal. 
\section{Toward AV 2.0: Unifying vision, language, and action within Embodied AI for safe and explainable end-to-end autonomous driving}
While the above subsections primarily describe the potential XAI approaches from the perspective of specific components, we also need to consider AVs' learning software as a holistic driving system. Three decades of research, starting with ALVINN in 1988 \cite{pomerleau1988alvinn} and further succeeding with the DARPA Grand Challenge \cite{thrun2006stanley}, have achieved significant milestones with traditional AI software. However, recent breakthroughs in Foundation Models in terms of LLMs and VLMs motivate a transition to next-generation AVs. This generation of AVs has been referred to as \textit{AV2.0} by industry professionals  \cite{jain2021autonomy,hawke2021reimagining, Waymo_e2e_2024}. The proposal is that the availability of integrated sensor suites, computational resources (i.e., GPU, TPU), and deep learning approaches can help AVs navigate via an end-to-end approach through adaptive learning, scaling, and generalization in complex driving environments. The ability to learn continually through interaction with the environment rather than relying on static datasets has resulted in the emergence of a new direction, labelled as ``Embodied AI'' \cite{duan2022survey, wang2024comprehensive}, and AV2.0 can move forward with such a learning approach. Effectively unifying vision, language, and action within Embodied AI can enable an autonomous car to navigate, interpret, and describe its high-level decisions in real-time. However, safety and explainability components of an end-end self-driving architecture  must overcome fundamental challenges in AI described below: \\
\textit{Safety}: The established guideline on core problems with AI safety \cite{amodei2016concrete} underscores five crucial considerations: avoiding negative side effects, avoiding reward hacking, scalable oversight, safe exploration, and robustness to distributional shift. We analyze the implications of these problems for end-to-end autonomous driving as follows:
\begin{itemize}[leftmargin=*]
    \item \textbf{Avoiding negative side effects}: 
    Autonomous driving is primarily associated with the ability of a self-driving car to avoid accidents and maintain a safe distance from stationary and dynamic objects along the planned motion trajectory. 
    However, the scope of the problem is not limited to this feature. Consider a scenario where an autonomous car interacts with another two vehicles, V1 and V2, at a specific moment. While aiming to make safe temporal decisions by itself, the autonomous car must also ensure that it does not implicitly enable  V1 and V2 to cause an accident at that road segment as a part of vehicle-to-vehicle (V2V) communication. According to \cite{amodei2016concrete}, a potential solution
    
     \begin{figure*}[htp!]
    \centering
    \includegraphics[width=1\textwidth]{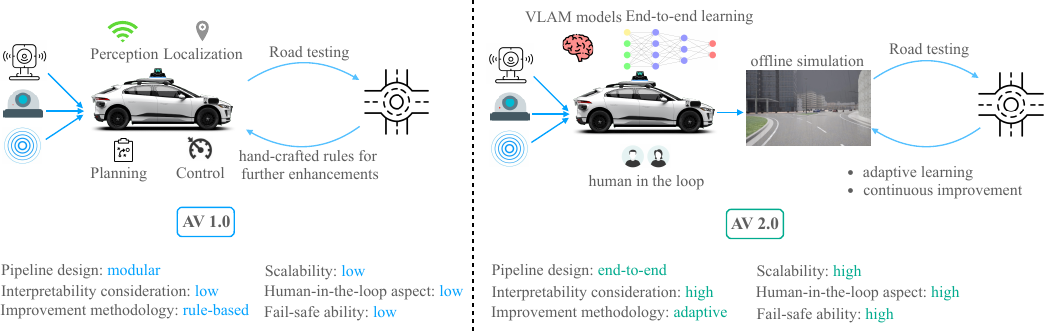}
    \caption{Our approach to AV2.0 vs AV1.0, and potential advantages of AV2.0 over AV1.0 in terms of its AI software stack, safety and explainability. The image of the vehicle has been taken from \cite{waymo_driver2024}.}
    \label{fig:AV2}
\end{figure*}
 to this problem could be to leverage cooperative Inverse RL \cite{hadfield2016cooperative}, where an autonomous system can cooperate with humans, and a human actor can always shut down the autonomous system in case such a system exhibits undesirable behavior. In the context of autonomous driving, this nuance can be related to an AV's communication with a human-operated vehicle or other remote operator monitoring an AV's overall driving safety. One of the prominent methods in this context is Sympathetic Cooperative Driving or SymCoDrive paradigm \cite{toghi2021cooperative}, which trains agents not only to achieve safe driving for themselves but also for human-controlled vehicles by promoting altruistic driving behavior in cooperative autonomous driving. As the deployment of AVs on roads is a gradual process, synergy with human-operated vehicles is a viable approach for socially aware and safe navigation.
    \item \textbf{Avoiding reward hacking}: Can we ensure that the end-to-end driving system does not shape its dynamic reward function according to what it sees in less dynamic environments and still apply that reward shaping while transitioning to highly dynamic environments? Particularly, as an embodied AI agent with adaptive learning and generalization ability in unseen environments, reward formulation must account for long horizons ahead and should not adjust its goals for short-term safe driving behavior. This topic has recently been well-investigated by Knox et al. \cite{knox2023reward}. They propose that flaws in reward shaping for RL-controlled autonomous driving can be identified by \textit{eight sanity checks}: unsafe reward shaping, potential mismatch between people and reward function's preferences, undesired risk tolerance via indifference points, learnable loopholes, missing attributes, redundant attributes, and trial-and-error reward design. The study discloses that such sanity checks can capture flaws in reward shaping for autonomous driving that can also exist in reward shaping for other tasks. 
    \item \textbf{Scalable oversight}: Can humans measure whether AVs perform at a human level or better in general in all driving situations, where in specific moments, evaluating the driving behavior of end-to-end driving may be difficult for humans due to some reasons. While being outside of human override, temporarily (i.e., refer to the Molly problem \cite{molly2020}), for various reasons, can we trust that AVs will behave safely at that moment? Amodei et al. \cite{amodei2016concrete} report that a potential solution to this problem may be semi-supervised RL: an agent can see its reward on a small subset of episodes or times steps. While rewards from all episodes are used to evaluate the agent's performance, the agent can only use that subset of rewards to optimize its performance under this setting.  
    \item \textbf{Safe exploration}: 
    Can an AV always make safe decisions when it has a binary choice of actions in a specific time interval? For example, an autonomous car may change its predefined route due to traffic congestion; however, the alternative route may have dangerous potholes or other damaged infrastructure that may lead to risky driving while attempting to save time on the trip. 
    \item \textbf{Robustness to distributional shift}: A well-known problem with AVs is making the transition from a simulation environment to real roads. For instance, RL-based end-to-end autonomous driving with an impressive performance in a simulation environment may not have the same performance when deployed to a physical autonomous car. Filos et al. \cite{filos2020can} have investigated this topic and proposed \textit{robust imitative planning}, a technique for epistemic uncertainty-aware planning. The key idea is that in case the model has great uncertainty in suggesting a safe course of action,
 the model can achieve sample-efficient online adaptation by querying the expert driver for feedback.
Through several experiments and state-of-the-art results, the authors also release  CARNOVEL, a benchmark for evaluating the robustness of driving agents with distribution shifts. Such a benchmark may be a significant part of a robust solution for addressing out-of-distribution scenarios.
\end{itemize}
These problems reflect a broad spectrum of potential safety issues with end-to-end AVs. However, we argue that the proposal misses yet another essential concept, namely \textit{fail-safe} ability. This concept has been investigated in some recent work \cite{magdici2016fail, xue2023fail, pek2020fail}; however, the recent proposals of the next-generation AVs \cite{jain2021autonomy,hawke2021reimagining, Waymo_e2e_2024} do not explicitly consider this functionality as an integral component of this technology. Human drivers often have a rest once they feel tired on long trips, and a short rest may help them feel mentally/physically better in the next phase of their driving. The same example can applied to AVs as well. Due to internal reasons (e.g., temporary system malfunction) or external factors (e.g., extremely adverse weather conditions), AVs may need to pause their trip temporarily and prevent further high-stakes consequences ahead. Such capability should not be considered a limitation of AVs; on the contrary, it is an optimal design strategy that foresees potential issues due to \textit{any} factors and makes AVs behave safely by directing them to ``have a short rest." \\
\textit{Explainability}: The reviewed studies in Section 4 show a significant milestone in the explainability of self-driving systems. However, there are still significant gaps and challenges to achieving accurate and timely explanations in all phases of trips. For instance, as of September 2023, it is reported that LINGO-1 exhibits roughly 60\% performance in its linguistic and VQA-based explanations compared to human-level performance \cite{wayve_lingo1_2023}.\\ 
Apart from informational content, another critical aspect deserving attention is the timing perspective of such explanations:  the lead time for emergent scenarios, perhaps using extensive scenario-based evaluations or case-based reasoning, must be engineered appropriately. Furthermore, a well-known problem with large pre-trained models, hallucinations, is another challenge in explanation delivery. Particularly, in QA models, the model must generate a response based on the joint question and scene-based semantics rather than being influenced by the question itself, such as in the case of adversarial QAs. We have recently performed an empirical study \cite{atakishiyev2024safety} on the latter and shown that even advanced VLMs can fail to detect the language bias in QA models and present incorrect explanations in case of human adversarial questions. This issue, in turn,  may damage user trust and can also have negative safety implications for self-driving. So, we argue that large pre-trained models' construction mechanism can be adjusted and regulated with common sense and human-defined concepts, as also posited by Kenny and Shah \cite{kenny2023pursuit}. Hence, designing robust QA models deserves more attention to enable meaningful and trustworthy dialogues between users and AVs. These features are key for achieving effective human-AI alignment  \cite{{sanneman2022situation, endsley2023supporting}}, trust \cite{ali2023explainable, diaz2023connecting, zhang2024critical}, and public acceptance \cite{omeiza2021explanations, zhang2024critical, ehsan2021operationalizing} with AV2.0. We describe our approach to AV2.0 and its difference from AV1.0 in Figure \ref{fig:AV2}.

\section{Conclusion}
We have presented a systematic overview of state-of-the-art investigations, emerging paradigms, and a future perspective of XAI approaches for autonomous driving. Insights from these studies reveal the existing gaps, and we have proposed a conceptual framework for explainable autonomous driving by incorporating missing pieces. The key idea is that AVs need to achieve regulatory-compliant operational safety and explainability in real-time decisions with their increasing automation ability. Together with a detailed overview of the state of the art in XAI-based autonomous driving, our work contributes as a \textit{cause-effect-solution} perspective. We elaborate on the notion of \textit{cause} by identifying current gaps, concerns, and a variety of issues specified while denoting \textit{effect} through the public reluctance on the use of autonomous driving at a broader level. We provide a \textit{solution} through the proposed framework and a set of promising XAI approaches for a future direction. This paper can benefit automotive researchers and practitioners in understanding the emerging paradigms and industrial trends in XAI approaches for autonomous driving and help achieve responsible, trustworthy and publicly acceptable next-generation AVs.

\section*{Acknowledgment}
We acknowledge support from the Alberta Machine Intelligence Institute
(Amii), from the Computing Science Department of the University of Alberta,
and the Natural Sciences and Engineering Research Council of Canada (NSERC). Shahin Atakishiyev also acknowledges support from the Ministry of Science and Education of the Republic of Azerbaijan.
 
\bibliography{IEEEabrv.bib, main.bib}{}

\begin{thebibliography}{100}
\providecommand{\url}[1]{#1}
\csname url@samestyle\endcsname
\providecommand{\newblock}{\relax}
\providecommand{\bibinfo}[2]{#2}
\providecommand{\BIBentrySTDinterwordspacing}{\spaceskip=0pt\relax}
\providecommand{\BIBentryALTinterwordstretchfactor}{4}
\providecommand{\BIBentryALTinterwordspacing}{\spaceskip=\fontdimen2\font plus
\BIBentryALTinterwordstretchfactor\fontdimen3\font minus \fontdimen4\font\relax}
\providecommand{\BIBforeignlanguage}[2]{{%
\expandafter\ifx\csname l@#1\endcsname\relax
\typeout{** WARNING: IEEEtran.bst: No hyphenation pattern has been}%
\typeout{** loaded for the language `#1'. Using the pattern for}%
\typeout{** the default language instead.}%
\else
\language=\csname l@#1\endcsname
\fi
#2}}
\providecommand{\BIBdecl}{\relax}
\BIBdecl

\bibitem{singh2015critical}
S.~Singh, ``Critical reasons for crashes investigated in the national motor vehicle crash causation survey,'' Tech. Rep., 2015.

\bibitem{lanctot2017accelerating}
R.~Lanctot \emph{et~al.}, ``Accelerating the future: The economic impact of the emerging passenger economy,'' \emph{Strategy Analytics}, vol.~5, 2017.

\bibitem{adadi2018peeking}
A.~Adadi and M.~Berrada, ``{Peeking Inside the Black-Box: A Survey on Explainable Artificial Intelligence (XAI)},'' \emph{IEEE Access}, vol.~6, pp. 52\,138--52\,160, 2018.

\bibitem{stanton2019models}
N.~A. Stanton, P.~M. Salmon, G.~H. Walker, and M.~Stanton, ``{Models and methods for collision analysis: a comparison study based on the Uber collision with a pedestrian},'' \emph{Safety Science}, vol. 120, pp. 117--128, 2019.

\bibitem{yurtsever2020survey}
E.~Yurtsever, J.~Lambert, A.~Carballo, and K.~Takeda, ``{A Survey of Autonomous Driving: Common Practices and Emerging Technologies},'' \emph{IEEE Access}, vol.~8, pp. 58\,443--58\,469, 2020.

\bibitem{board2020collision}
{NTS Board}, ``Collision between a sport utility vehicle operating with partial driving automation and a crash attenuator {Mountain View}, {California.}'' { Accessed online} on February 10, 2023.

\bibitem{ali2023explainable}
S.~Ali, T.~Abuhmed, S.~El-Sappagh, K.~Muhammad, J.~M. Alonso-Moral, R.~Confalonieri, R.~Guidotti, J.~Del~Ser, N.~D{\'\i}az-Rodr{\'\i}guez, and F.~Herrera, ``{Explainable Artificial Intelligence (XAI): What we know and what is left to attain Trustworthy Artificial Intelligence},'' \emph{Information Fusion}, vol.~99, p. 101805, 2023.

\bibitem{dong2023did}
J.~Dong, S.~Chen, M.~Miralinaghi, T.~Chen, P.~Li, and S.~Labi, ``{Why did the AI make that decision? Towards an explainable artificial intelligence (XAI) for autonomous driving systems},'' \emph{Transportation Research Part C: Emerging Technologies}, vol. 156, p. 104358, 2023.

\bibitem{saeed2023explainable}
W.~Saeed and C.~Omlin, ``{Explainable AI (XAI): A systematic meta-survey of current challenges and future opportunities},'' \emph{Knowledge-Based Systems}, vol. 263, p. 110273, 2023.

\bibitem{daimler2017}
{Daimler media}. {Autonomous concept car smart vision EQ fortwo: Welcome to the future of car sharing}. (Accessed on March 10, 2024).

\bibitem{badue2021self}
C.~Badue, R.~Guidolini, R.~V. Carneiro, P.~Azevedo, V.~B. Cardoso, A.~Forechi, L.~Jesus, R.~Berriel, T.~M. Paixao, F.~Mutz \emph{et~al.}, ``Self-driving cars: A survey,'' \emph{Expert Systems with Applications}, vol. 165, p. 113816, 2021.

\bibitem{campbell2018sensor}
S.~Campbell, N.~O'Mahony, L.~Krpalcova, D.~Riordan, J.~Walsh, A.~Murphy, and C.~Ryan, ``Sensor technology in autonomous vehicles: A review,'' in \emph{2018 29th Irish Signals and Systems Conference (ISSC)}.\hskip 1em plus 0.5em minus 0.4em\relax IEEE, 2018, pp. 1--4.

\bibitem{yeong2021sensor}
D.~J. Yeong, G.~Velasco-Hernandez, J.~Barry, J.~Walsh \emph{et~al.}, ``{Sensor and Sensor Fusion Technology in Autonomous Vehicles: A Review},'' \emph{Sensors}, vol.~21, no.~6, p. 2140, 2021.

\bibitem{pomerleau1988alvinn}
D.~A. Pomerleau, ``{ALVINN: An Autonomous Land Vehicle in a Neural Network},'' \emph{Advances in Neural Information Processing Systems}, vol.~1, 1988.

\bibitem{Shuttleworth}
J.~Shuttleworth, ``{SAE} standard news: J3016 automated-driving graphic update,'' \url{https://www.sae.org/news/2019/01/sae-updates-j3016-automated-driving-graphic}, 2019, {Accessed online} on August 16, 2021.

\bibitem{chen2023end}
L.~Chen, P.~Wu, K.~Chitta, B.~Jaeger, A.~Geiger, and H.~Li, ``{End-to-end Autonomous Driving: Challenges and Frontiers},'' \emph{arXiv preprint arXiv:2306.16927}, 2023.

\bibitem{hu2023planning}
Y.~Hu, J.~Yang, L.~Chen, K.~Li, C.~Sima, X.~Zhu, S.~Chai, S.~Du, T.~Lin, W.~Wang \emph{et~al.}, ``{Planning-oriented Autonomous Driving},'' in \emph{Proceedings of the IEEE/CVF Conference on Computer Vision and Pattern Recognition}, 2023, pp. 17\,853--17\,862.

\bibitem{araluce2024leveraging}
J.~Araluce, L.~M. Bergasa, M.~Oca{\~n}a, {\'A}.~Llamazares, and E.~L{\'o}pez-Guill{\'e}n, ``{Leveraging Driver Attention for an End-to-End Explainable Decision-Making From Frontal Images},'' \emph{IEEE Transactions on Intelligent Transportation Systems}, 2024.

\bibitem{tampuu2020survey}
A.~Tampuu, T.~Matiisen, M.~Semikin, D.~Fishman, and N.~Muhammad, ``{A Survey of End-to-End Driving: Architectures and Training Methods},'' \emph{IEEE Transactions on Neural Networks and Learning Systems}, vol.~33, no.~4, pp. 1364--1384, 2020.

\bibitem{hu2022st}
S.~Hu, L.~Chen, P.~Wu, H.~Li, J.~Yan, and D.~Tao, ``{ST-P3: End-to-End Vision-Based Autonomous Driving via Spatial-Temporal Feature Learning},'' in \emph{European Conference on Computer Vision}.\hskip 1em plus 0.5em minus 0.4em\relax Springer, 2022, pp. 533--549.

\bibitem{jing2022inaction}
T.~Jing, H.~Xia, R.~Tian, H.~Ding, X.~Luo, J.~Domeyer, R.~Sherony, and Z.~Ding, ``{InAction: Interpretable action decision making for autonomous driving},'' in \emph{Proceedings of the 2022 European Conference on Computer Vision}.\hskip 1em plus 0.5em minus 0.4em\relax Springer, 2022, pp. 370--387.

\bibitem{muller2019comparing}
J.~M. M{\"u}ller, ``Comparing technology acceptance for autonomous vehicles, battery electric vehicles, and car sharing—a study across {Europe}, {China}, and {North America},'' \emph{Sustainability}, vol.~11, no.~16, p. 4333, 2019.

\bibitem{fleetwood2017public}
J.~Fleetwood, ``Public health, ethics, and autonomous vehicles,'' \emph{American Journal of Public Health}, vol. 107, no.~4, pp. 532--537, 2017.

\bibitem{foot1967problem}
P.~Foot, ``The problem of abortion and the doctrine of the double effect,'' \emph{Oxford review}, vol.~5, 1967.

\bibitem{jarvis1985trolley}
J.~Jarvis~Thomson, ``The trolley problem,'' \emph{Yale Law Journal}, vol.~94, no.~6, p.~5, 1985.

\bibitem{awad2018moral}
E.~Awad, S.~Dsouza, R.~Kim, J.~Schulz, J.~Henrich, A.~Shariff, J.-F. Bonnefon, and I.~Rahwan, ``{The Moral Machine experiment },'' \emph{Nature}, vol. 563, no. 7729, pp. 59--64, 2018.

\bibitem{lundgren2020safety}
B.~Lundgren, ``Safety requirements vs. crashing ethically: what matters most for policies on autonomous vehicles,'' \emph{AI \& SOCIETY}, pp. 1--11, 2020.

\bibitem{harris2020immoral}
J.~Harris, ``The immoral machine,'' \emph{Cambridge Quarterly of Healthcare Ethics}, vol.~29, no.~1, pp. 71--79, 2020.

\bibitem{burton2020mind}
S.~Burton, I.~Habli, T.~Lawton, J.~McDermid, P.~Morgan, and Z.~Porter, ``Mind the gaps: Assuring the safety of autonomous systems from an engineering, ethical, and legal perspective,'' \emph{Artificial Intelligence}, vol. 279, p. 103201, 2020.

\bibitem{sohrabi2020impacts}
S.~Sohrabi, H.~Khreis, and D.~Lord, ``{Impacts of Autonomous Vehicles on Public Health: A Conceptual Model and Policy Recommendations},'' \emph{Sustainable Cities and Society}, vol.~63, p. 102457, 2020.

\bibitem{martinho2021ethical}
A.~Martinho, N.~Herber, M.~Kroesen, and C.~Chorus, ``Ethical issues in focus by the autonomous vehicles industry,'' \emph{Transport Reviews}, pp. 1--22, 2021.

\bibitem{regulation2016}
GDPR, ``Regulation eu 2016/679 of the european parliament and of the council of 27 april 2016,'' \emph{Official Journal of the European Union}, 2016.

\bibitem{eustrategy2019}
{The High-Level Expert Group on AI at the European Commission}, ``{Ethics guidelines for trustworthy AI },'' \url{https://digital-strategy.ec.europa.eu/en/library/ethics-guidelines-trustworthy-ai#:~:text=On%208%20April%202019%2C%20the,received%20through%20an%20open%20consultation.}, 2019, accessed on March 11, 2024.

\bibitem{nacto2016}
NACTO, ``{NACTO} policy statement on automated vehicles,'' \url{https://nacto.org/wp-content/uploads/2016/06/NACTO-Policy-Automated-Vehicles-201606.pdf}, 2016, accessed March 10, 2024.

\bibitem{national2016federal}
{National Highway Traffic Safety Administration}, \emph{Federal automated vehicles policy: Accelerating the next revolution in roadway safety}.\hskip 1em plus 0.5em minus 0.4em\relax US Department of Transportation, 2016.

\bibitem{NHTSA2022}
{Department of Transportation}, ``Occupant protection for vehicles with automated driving systems,'' \url{https://www.nhtsa.gov/sites/nhtsa.gov/files/2022-03/Final-Rule-Occupant-Protection-Amendment-Automated-Vehicles.pdf}, 2022.

\bibitem{guideleinescanada}
{Transport Canada}, ``Guidelines for testing automated driving systems in {Canada},'' \emph{The Ministry of Transportation of Canada}, 2021.

\bibitem{germany2021}
{Bundesanzeiger Verlag Board}, ``Act amending the road traffic act and the compulsory insurance act (autonomous driving act),'' 2021.

\bibitem{UK2021}
{Department for Transport team}, ``Safe use of automated lane keeping system (alks) summary of responses and next steps,'' 2021.

\bibitem{austroads2017guidelines}
Austroads, \emph{Guidelines for trials of automated vehicles in Australia}.\hskip 1em plus 0.5em minus 0.4em\relax National Transport Commission, 2017.

\bibitem{japan2017}
{Advanced information and telecommunications network society}, \emph{Outline of systematic preparations related to autonomous driving}.\hskip 1em plus 0.5em minus 0.4em\relax {The Government of Japan}, 2017.

\bibitem{ISO21448}
\BIBentryALTinterwordspacing
{ISO 21448 Committee}. {Road vehicles — Safety of the intended functionality}. (Accessed on February 10, 2023). [Online]. Available: \url{https://www.iso.org/standard/70939.html}
\BIBentrySTDinterwordspacing

\bibitem{ISO26262}
\BIBentryALTinterwordspacing
{ISO 26262 Committee}. {Road vehicles — Functional safety}. (Accessed on February 10, 2023). [Online]. Available: \url{https://www.iso.org/standard/68383.html}
\BIBentrySTDinterwordspacing

\bibitem{apexdocs}
\BIBentryALTinterwordspacing
{Apex.AI Blog}. {An overview of taxonomy, legislation, regulations, and standards for automated mobility}. (Accessed on April 8, 2024). [Online]. Available: \url{https://www.apex.ai/post/legislation-standards-taxonomy-overview}
\BIBentrySTDinterwordspacing

\bibitem{omeiza2021explanations}
D.~Omeiza, H.~Webb, M.~Jirotka, and L.~Kunze, ``{Explanations in Autonomous Driving: A Survey},'' \emph{IEEE Transactions on Intelligent Transportation Systems}, 2021.

\bibitem{ehsan2020human}
U.~Ehsan and M.~O. Riedl, ``Human-centered explainable ai: towards a reflective sociotechnical approach,'' in \emph{International Conference on Human-Computer Interaction}.\hskip 1em plus 0.5em minus 0.4em\relax Springer, 2020, pp. 449--466.

\bibitem{dhanorkar2021needs}
S.~Dhanorkar, C.~T. Wolf, K.~Qian, A.~Xu, L.~Popa, and Y.~Li, ``Who needs to know what, when?: Broadening the explainable ai (xai) design space by looking at explanations across the ai lifecycle,'' in \emph{Designing Interactive Systems Conference 2021}, 2021, pp. 1591--1602.

\bibitem{lewis1986causal}
D.~Lewis, ``Causal explanation,'' 1986.

\bibitem{miller2019explanation}
T.~Miller, ``Explanation in artificial intelligence: Insights from the social sciences,'' \emph{Artificial Intelligence}, vol. 267, pp. 1--38, 2019.

\bibitem{pearl2009causality}
J.~Pearl, \emph{Causality}.\hskip 1em plus 0.5em minus 0.4em\relax Cambridge University Press, 2009.

\bibitem{design2016vision}
{IEEE Global Initiative}, ``A vision for prioritizing human well-being with artificial intelligence and autonomous systems,'' \emph{IEEE Glob Initiat Ethical Considerations Artif Intell Auton Syst}, vol.~13, 2016.

\bibitem{israelsen2019dave}
B.~W. Israelsen and N.~R. Ahmed, ``{“Dave... I can assure you... that it’s going to be all right...” A definition, case for, and survey of algorithmic assurances in human-autonomy trust relationships},'' \emph{ACM Computing Surveys (CSUR)}, vol.~51, no.~6, pp. 1--37, 2019.

\bibitem{Mercedes2022}
{David Mullen}, ``{Mercedes to accept legal responsibility for accidents involving self-driving cars},'' 2022.

\bibitem{arfini2023design}
S.~Arfini, P.~Bellani, A.~Picardi, M.~Yan, F.~Fossa, and G.~Caruso, ``Design for inclusivity in driving automation: Theoretical and practical challenges to human-machine interactions and interface design,'' in \emph{Connected and Automated Vehicles: Integrating Engineering and Ethics}.\hskip 1em plus 0.5em minus 0.4em\relax Springer, 2023, pp. 63--85.

\bibitem{langley2019varieties}
P.~Langley, ``Varieties of explainable agency,'' in \emph{ICAPS Workshop on Explainable AI Planning (XAIP)}, 2019.

\bibitem{mittelstadt2019explaining}
B.~Mittelstadt, C.~Russell, and S.~Wachter, ``Explaining explanations in ai,'' in \emph{Proceedings of the Conference on Fairness, Accountability, and Transparency}, 2019, pp. 279--288.

\bibitem{arrieta2020explainable}
A.~B. Arrieta, N.~D{\'\i}az-Rodr{\'\i}guez, J.~Del~Ser, A.~Bennetot, S.~Tabik, A.~Barbado, S.~Garc{\'\i}a, S.~Gil-L{\'o}pez, D.~Molina, R.~Benjamins \emph{et~al.}, ``{Explainable Artificial Intelligence (XAI): Concepts, taxonomies, opportunities and challenges toward responsible AI},'' \emph{Information Fusion}, vol.~58, pp. 82--115, 2020.

\bibitem{zablocki2021explainability}
{\'E}.~Zablocki, H.~Ben-Younes, P.~P{\'e}rez, and M.~Cord, ``Explainability of deep vision-based autonomous driving systems: Review and challenges,'' \emph{International Journal of Computer Vision}, vol. 130, no.~10, pp. 2425--2452, 2022.

\bibitem{herlocker2000explaining}
J.~L. Herlocker, J.~A. Konstan, and J.~Riedl, ``Explaining collaborative filtering recommendations,'' in \emph{Proceedings of the 2000 ACM Conference on Computer Supported Cooperative Work}, 2000, pp. 241--250.

\bibitem{roth2004explanations}
T.~R. Roth-Berghofer, ``Explanations and case-based reasoning: Foundational issues,'' in \emph{European Conference on Case-Based Reasoning}.\hskip 1em plus 0.5em minus 0.4em\relax Springer, 2004, pp. 389--403.

\bibitem{lim2009assessing}
B.~Y. Lim and A.~K. Dey, ``Assessing demand for intelligibility in context-aware applications,'' in \emph{Proceedings of the 11th International Conference on Ubiquitous Computing}, 2009, pp. 195--204.

\bibitem{liao2020questioning}
Q.~V. Liao, D.~Gruen, and S.~Miller, ``{Questioning the AI: Informing Design Practices for Explainable AI User Experiences},'' in \emph{Proceedings of the 2020 CHI Conference on Human Factors in Computing Systems}, 2020, pp. 1--15.

\bibitem{wang2019designing}
D.~Wang, Q.~Yang, A.~Abdul, and B.~Y. Lim, ``{Designing Theory-Driven User-Centric Explainable AI},'' in \emph{Proceedings of the 2019 CHI Conference on Human Factors in Computing Systems}, 2019, pp. 1--15.

\bibitem{naujoks2018use}
F.~Naujoks, S.~Hergeth, K.~Wiedemann, N.~Sch{\"o}mig, and A.~Keinath, ``Use cases for assessing, testing, and validating the human machine interface of automated driving systems,'' in \emph{Proceedings of the Human Factors and Ergonomics Society Annual Meeting}, vol.~62, no.~1.\hskip 1em plus 0.5em minus 0.4em\relax Sage Publications Sage CA: Los Angeles, CA, 2018, pp. 1873--1877.

\bibitem{schneider2021increasing}
T.~Schneider, S.~Ghellal, S.~Love, and A.~R. Gerlicher, ``Increasing the user experience in autonomous driving through different feedback modalities,'' in \emph{26th International Conference on Intelligent User Interfaces}, 2021, pp. 7--10.

\bibitem{bojarski2016visualbackprop}
M.~Bojarski, A.~Choromanska, K.~Choromanski, B.~Firner, L.~Jackel, U.~Muller, and K.~Zieba, ``{VisualBackProp: efficient visualization of CNNs},'' \emph{arXiv preprint arXiv:1611.05418}, vol.~2, 2016.

\bibitem{kim2017interpretable}
J.~Kim and J.~Canny, ``Interpretable learning for self-driving cars by visualizing causal attention,'' in \emph{2017 IEEE International Conference on Computer Vision (ICCV)}.\hskip 1em plus 0.5em minus 0.4em\relax IEEE, 2017, pp. 2961--2969.

\bibitem{kim2018textual}
J.~Kim, A.~Rohrbach, T.~Darrell, J.~Canny, and Z.~Akata, ``Textual explanations for self-driving vehicles,'' in \emph{Proceedings of the European Conference on Computer Vision (ECCV)}, 2018, pp. 563--578.

\bibitem{hofmarcher2019visual}
M.~Hofmarcher, T.~Unterthiner, J.~Arjona-Medina, G.~Klambauer, S.~Hochreiter, and B.~Nessler, ``Visual scene understanding for autonomous driving using semantic segmentation,'' in \emph{Explainable AI: Interpreting, Explaining and Visualizing Deep Learning}.\hskip 1em plus 0.5em minus 0.4em\relax Springer, 2019, pp. 285--296.

\bibitem{zeng2019end}
W.~Zeng, W.~Luo, S.~Suo, A.~Sadat, B.~Yang, S.~Casas, and R.~Urtasun, ``End-to-end interpretable neural motion planner,'' in \emph{Proceedings of the IEEE/CVF Conference on Computer Vision and Pattern Recognition}, 2019, pp. 8660--8669.

\bibitem{hu2019multi}
Y.~Hu, W.~Zhan, L.~Sun, and M.~Tomizuka, ``Multi-modal probabilistic prediction of interactive behavior via an interpretable model,'' in \emph{2019 IEEE Intelligent Vehicles Symposium (IV)}.\hskip 1em plus 0.5em minus 0.4em\relax IEEE, 2019, pp. 557--563.

\bibitem{xu2020explainable}
Y.~Xu, X.~Yang, L.~Gong, H.-C. Lin, T.-Y. Wu, Y.~Li, and N.~Vasconcelos, ``Explainable object-induced action decision for autonomous vehicles,'' in \emph{Proceedings of the IEEE/CVF Conference on Computer Vision and Pattern Recognition}, 2020, pp. 9523--9532.

\bibitem{kim2020advisable}
J.~Kim, S.~Moon, A.~Rohrbach, T.~Darrell, and J.~Canny, ``Advisable learning for self-driving vehicles by internalizing observation-to-action rules,'' in \emph{Proceedings of the IEEE/CVF Conference on Computer Vision and Pattern Recognition}, 2020, pp. 9661--9670.

\bibitem{li2020make}
C.~Li, S.~H. Chan, and Y.-T. Chen, ``{Who Make Drivers Stop? Towards Driver-centric Risk Assessment: Risk Object Identification via Causal Inference},'' in \emph{2020 IEEE/RSJ International Conference on Intelligent Robots and Systems (IROS)}.\hskip 1em plus 0.5em minus 0.4em\relax IEEE, 2020, pp. 10\,711--10\,718.

\bibitem{casas2021mp3}
S.~Casas, A.~Sadat, and R.~Urtasun, ``{MP3: A Unified Model to Map, Perceive, Predict and Plan},'' in \emph{Proceedings of the IEEE/CVF Conference on Computer Vision and Pattern Recognition}, 2021, pp. 14\,403--14\,412.

\bibitem{kim2021toward}
J.~Kim, A.~Rohrbach, Z.~Akata, S.~Moon, T.~Misu, Y.-T. Chen, T.~Darrell, and J.~Canny, ``Toward explainable and advisable model for self-driving cars,'' \emph{Applied AI Letters}, p. e56, 2021.

\bibitem{wang2021human}
C.~Wang, T.~H. Weisswange, M.~Krueger, and C.~B. Wiebel-Herboth, ``{Human-Vehicle Cooperation on Prediction-Level: Enhancing Automated Driving with Human Foresight},'' in \emph{2021 IEEE Intelligent Vehicles Symposium Workshops (IV Workshops)}.\hskip 1em plus 0.5em minus 0.4em\relax IEEE, 2021, pp. 25--30.

\bibitem{chitta2021neat}
K.~Chitta, A.~Prakash, and A.~Geiger, ``{NEAT: Neural Attention Fields for End-to-End Autonomous Driving},'' in \emph{Proceedings of the IEEE/CVF International Conference on Computer Vision}, 2021, pp. 15\,793--15\,803.

\bibitem{dong2021image}
J.~Dong, S.~Chen, S.~Zong, T.~Chen, and S.~Labi, ``{Image Transformer for Explainable Autonomous Driving System},'' in \emph{2021 IEEE International Intelligent Transportation Systems Conference (ITSC)}.\hskip 1em plus 0.5em minus 0.4em\relax IEEE, 2021, pp. 2732--2737.

\bibitem{hanna2021interpretable}
J.~P. Hanna, A.~Rahman, E.~Fosong, F.~Eiras, M.~Dobre, J.~Redford, S.~Ramamoorthy, and S.~V. Albrecht, ``{Interpretable Goal Recognition in the Presence of Occluded Factors for Autonomous Vehicles},'' in \emph{2021 IEEE/RSJ International Conference on Intelligent Robots and Systems (IROS)}.\hskip 1em plus 0.5em minus 0.4em\relax IEEE, 2021, pp. 7044--7051.

\bibitem{mankodiya2021xai}
H.~Mankodiya, M.~S. Obaidat, R.~Gupta, and S.~Tanwar, ``{XAI-AV: Explainable Artificial Intelligence for Trust Management in Autonomous Vehicles},'' in \emph{2021 International Conference on Communications, Computing, Cybersecurity, and Informatics (CCCI)}.\hskip 1em plus 0.5em minus 0.4em\relax IEEE, 2021, pp. 1--5.

\bibitem{madhav2022explainable}
A.~S. Madhav and A.~K. Tyagi, ``{Explainable Artificial Intelligence (XAI): Connecting Artificial Decision-Making and Human Trust in Autonomous Vehicles},'' in \emph{Proceedings of Third International Conference on Computing, Communications, and Cyber-Security: IC4S 2021}.\hskip 1em plus 0.5em minus 0.4em\relax Springer, 2022, pp. 123--136.

\bibitem{jacob2022steex}
P.~Jacob, {\'E}.~Zablocki, H.~Ben-Younes, M.~Chen, P.~P{\'e}rez, and M.~Cord, ``{STEEX: Steering Counterfactual Explanations with Semantics},'' in \emph{Computer Vision--ECCV 2022: 17th European Conference, Tel Aviv, Israel, October 23--27, 2022, Proceedings, Part XII}.\hskip 1em plus 0.5em minus 0.4em\relax Springer, 2022, pp. 387--403.

\bibitem{zhang2022attention}
Z.~Zhang, R.~Tian, R.~Sherony, J.~Domeyer, and Z.~Ding, ``{Attention-Based Interrelation Modeling for Explainable Automated Driving},'' \emph{IEEE Transactions on Intelligent Vehicles}, 2022.

\bibitem{kolekar2022explainable}
S.~Kolekar, S.~Gite, B.~Pradhan, and A.~Alamri, ``{Explainable AI in scene understanding for autonomous vehicles in unstructured traffic environments on Indian roads using the inception U-Net Model with Grad-CAM visualization},'' \emph{sensors}, vol.~22, no.~24, p. 9677, 2022.

\bibitem{zemni2023octet}
M.~Zemni, M.~Chen, {\'E}.~Zablocki, H.~Ben-Younes, P.~P{\'e}rez, and M.~Cord, ``{OCTET: Object-aware Counterfactual Explanations},'' in \emph{Proceedings of the IEEE/CVF Conference on Computer Vision and Pattern Recognition}, 2023, pp. 15\,062--15\,071.

\bibitem{itkina2023interpretable}
M.~Itkina and M.~Kochenderfer, ``{Interpretable Self-Aware Neural Networks for Robust Trajectory Prediction},'' in \emph{Conference on Robot Learning}.\hskip 1em plus 0.5em minus 0.4em\relax PMLR, 2023, pp. 606--617.

\bibitem{feng2023nle}
Y.~Feng, W.~Hua, and Y.~Sun, ``Nle-dm: Natural-language explanations for decision making of autonomous driving based on semantic scene understanding,'' \emph{IEEE Transactions on Intelligent Transportation Systems}, 2023.

\bibitem{hu2023holistic}
H.~Hu, Q.~Wang, Z.~Zhang, Z.~Li, and Z.~Gao, ``Holistic transformer: A joint neural network for trajectory prediction and decision-making of autonomous vehicles,'' \emph{Pattern Recognition}, vol. 141, p. 109592, 2023.

\bibitem{atakishiyev2023explaining}
S.~Atakishiyev, M.~Salameh, H.~Babiker, and R.~Goebel, ``{Explaining Autonomous Driving Actions with Visual Question Answering},'' in \emph{2023 IEEE 26th International Conference on Intelligent Transportation Systems (ITSC)}.\hskip 1em plus 0.5em minus 0.4em\relax IEEE, 2023, pp. 1207--1214.

\bibitem{echterhoff2024driving}
J.~Echterhoff, A.~Yan, K.~Han, A.~Abdelraouf, R.~Gupta, and J.~McAuley, ``Driving through the concept gridlock: Unraveling explainability bottlenecks in automated driving,'' in \emph{Proceedings of the IEEE/CVF Winter Conference on Applications of Computer Vision}, 2024, pp. 7346--7355.

\bibitem{feng2024polarpoint}
Y.~Feng and Y.~Sun, ``{PolarPoint-BEV: Bird-eye-view Perception in Polar Points for Explainable End-to-end Autonomous Driving},'' \emph{IEEE Transactions on Intelligent Vehicles}, 2024.

\bibitem{kuznietsov2024explainable}
A.~Kuznietsov, B.~Gyevnar, C.~Wang, S.~Peters, and S.~V. Albrecht, ``{Explainable AI for Safe and Trustworthy Autonomous Driving: A Systematic Review},'' \emph{arXiv preprint arXiv:2402.10086}, 2024.

\bibitem{zeiler2014visualizing}
M.~D. Zeiler and R.~Fergus, ``Visualizing and understanding convolutional networks,'' in \emph{European Conference on Computer Vision}.\hskip 1em plus 0.5em minus 0.4em\relax Springer, 2014, pp. 818--833.

\bibitem{hendricks2016generating}
L.~A. Hendricks, Z.~Akata, M.~Rohrbach, J.~Donahue, B.~Schiele, and T.~Darrell, ``{Generating Visual Explanations},'' in \emph{{European Conference on Computer Vision}}.\hskip 1em plus 0.5em minus 0.4em\relax Springer, 2016, pp. 3--19.

\bibitem{zhou2016learning}
B.~Zhou, A.~Khosla, A.~Lapedriza, A.~Oliva, and A.~Torralba, ``Learning deep features for discriminative localization,'' in \emph{Proceedings of the IEEE Conference on Computer Vision and Pattern Recognition}, 2016, pp. 2921--2929.

\bibitem{selvaraju2017grad}
R.~R. Selvaraju, M.~Cogswell, A.~Das, R.~Vedantam, D.~Parikh, and D.~Batra, ``Grad-cam: Visual explanations from deep networks via gradient-based localization,'' in \emph{Proceedings of the IEEE International Conference on Computer Vision}, 2017, pp. 618--626.

\bibitem{springenberg2014striving}
J.~T. Springenberg, A.~Dosovitskiy, T.~Brox, and M.~Riedmiller, ``Striving for simplicity: The all convolutional net,'' \emph{arXiv preprint arXiv:1412.6806}, 2014.

\bibitem{samek2016interpreting}
W.~Samek, G.~Montavon, A.~Binder, S.~Lapuschkin, and K.-R. M{\"u}ller, ``Interpreting the predictions of complex ml models by layer-wise relevance propagation,'' \emph{arXiv preprint arXiv:1611.08191}, 2016.

\bibitem{lapuschkin2019unmasking}
S.~Lapuschkin, S.~W{\"a}ldchen, A.~Binder, G.~Montavon, W.~Samek, and K.-R. M{\"u}ller, ``Unmasking clever hans predictors and assessing what machines really learn,'' \emph{Nature Communications}, vol.~10, no.~1, pp. 1--8, 2019.

\bibitem{shrikumar2017learning}
A.~Shrikumar, P.~Greenside, and A.~Kundaje, ``Learning important features through propagating activation differences,'' in \emph{International Conference on Machine Learning}.\hskip 1em plus 0.5em minus 0.4em\relax PMLR, 2017, pp. 3145--3153.

\bibitem{babiker2017introduction}
H.~K.~B. Babiker and R.~Goebel, ``{An Introduction to Deep Visual Explanation},'' \emph{31st Neural Information Processing Systems Conference (NIPS), Long Beach, CA, USA}, 2017.

\bibitem{babiker2017using}
H.~Babiker and R.~Goebel, ``{Using KL-divergence to focus Deep Visual Explanation},'' \emph{31st Neural Information Processing Systems Conference (NIPS), Interpretable ML Symposium. Long Beach, CA, USA}, 2017.

\bibitem{cordts2016cityscapes}
M.~Cordts, M.~Omran, S.~Ramos, T.~Rehfeld, M.~Enzweiler, R.~Benenson, U.~Franke, S.~Roth, and B.~Schiele, ``The cityscapes dataset for semantic urban scene understanding,'' in \emph{Proceedings of the IEEE Conference on Computer Vision and Pattern Recognition}, 2016, pp. 3213--3223.

\bibitem{CommaAIdataset}
{Comma.AI}, ``Public driving dataset,'' \url{https://github.com/commaai/research}, {Accessed online} on Apr 1, 2024.

\bibitem{Udacitydataset}
{Udacity}, ``Public driving dataset,'' \url{https://public.roboflow.com/object-detection/self-driving-car}, {Accessed online} on Apr 6, 2022.

\bibitem{yu2020bdd100k}
F.~Yu, H.~Chen, X.~Wang, W.~Xian, Y.~Chen, F.~Liu, V.~Madhavan, and T.~Darrell, ``{BDD100K: A Diverse Driving Dataset for Heterogeneous Multitask Learning},'' in \emph{Proceedings of the IEEE/CVF Conference on Computer Vision and Pattern Recognition}, 2020, pp. 2636--2645.

\bibitem{hochreiter1997long}
S.~Hochreiter and J.~Schmidhuber, ``Long short-term memory,'' \emph{Neural Computation}, vol.~9, no.~8, pp. 1735--1780, 1997.

\bibitem{simonyan2015vgg}
K.~Simonyan and A.~Zisserman, ``{Very Deep Convolutional Networks for Large-Scale Image Recognition},'' \emph{International Conference on Learning Representations}, 2015.

\bibitem{heuillet2021explainability}
A.~Heuillet, F.~Couthouis, and N.~D{\'\i}az-Rodr{\'\i}guez, ``Explainability in deep reinforcement learning,'' \emph{Knowledge-Based Systems}, vol. 214, p. 106685, 2021.

\bibitem{milani2023explainable}
S.~Milani, N.~Topin, M.~Veloso, and F.~Fang, ``{Explainable Reinforcement Learning: A Survey and Comparative Review},'' \emph{ACM Computing Surveys}, 2023.

\bibitem{bekkemoen2024explainable}
Y.~Bekkemoen, ``{Explainable reinforcement learning (XRL): a systematic literature review and taxonomy},'' \emph{Machine Learning}, vol. 113, no.~1, pp. 355--441, 2024.

\bibitem{pan2019semantic}
X.~Pan, X.~Chen, Q.~Cai, J.~Canny, and F.~Yu, ``{Semantic Predictive Control for Explainable and Efficient Policy Learning},'' in \emph{2019 International Conference on Robotics and Automation (ICRA)}.\hskip 1em plus 0.5em minus 0.4em\relax IEEE, 2019, pp. 3203--3209.

\bibitem{chen2021interpretable}
J.~Chen, S.~E. Li, and M.~Tomizuka, ``{Interpretable End-to-End Urban Autonomous Driving With Latent Deep Reinforcement Learning},'' \emph{IEEE Transactions on Intelligent Transportation Systems}, 2021.

\bibitem{dosovitskiy2017carla}
A.~Dosovitskiy, G.~Ros, F.~Codevilla, A.~Lopez, and V.~Koltun, ``{CARLA: An Open Urban Driving Simulator },'' in \emph{Conference on Robot Learning}.\hskip 1em plus 0.5em minus 0.4em\relax PMLR, 2017, pp. 1--16.

\bibitem{wang2021learning}
H.~Wang, P.~Cai, Y.~Sun, L.~Wang, and M.~Liu, ``{Learning Interpretable End-to-End Vision-Based Motion Planning for Autonomous Driving with Optical Flow Distillation},'' in \emph{2021 IEEE International Conference on Robotics and Automation (ICRA)}.\hskip 1em plus 0.5em minus 0.4em\relax IEEE, 2021, pp. 13\,731--13\,737.

\bibitem{caesar2020nuscenes}
H.~Caesar, V.~Bankiti, A.~H. Lang, S.~Vora, V.~E. Liong, Q.~Xu, A.~Krishnan, Y.~Pan, G.~Baldan, and O.~Beijbom, ``{nuScenes: A Multimodal Dataset for Autonomous Driving},'' in \emph{Proceedings of the IEEE/CVF Conference on Computer Vision and Pattern Recognition}, 2020, pp. 11\,621--11\,631.

\bibitem{wang2021uncovering}
H.~Wang, W.~Wang, S.~Yuan, and X.~Li, ``{Uncovering Interpretable Internal States of Merging Tasks at Highway on-Ramps for Autonomous Driving Decision-Making},'' \emph{IEEE Transactions on Automation Science and Engineering}, 2021.

\bibitem{bansalchauffeurnet}
M.~Bansal, A.~Krizhevsky, and A.~Ogale, ``{ChauffeurNet: Learning to Drive by Imitating the Best and Synthesizing the Worst},'' \emph{Robotics: Science and Systems}, 2018.

\bibitem{cultrera2020explaining}
L.~Cultrera, L.~Seidenari, F.~Becattini, P.~Pala, and A.~Del~Bimbo, ``{Explaining Autonomous Driving by Learning End-to-End Visual Attention},'' in \emph{Proceedings of the IEEE/CVF Conference on Computer Vision and Pattern Recognition Workshops}, 2020, pp. 340--341.

\bibitem{schmidt2021can}
L.~M. Schmidt, G.~Kontes, A.~Plinge, and C.~Mutschler, ``{Can You Trust Your Autonomous Car? Interpretable and Verifiably Safe Reinforcement Learning},'' in \emph{2021 IEEE Intelligent Vehicles Symposium (IV)}.\hskip 1em plus 0.5em minus 0.4em\relax IEEE, 2021, pp. 171--178.

\bibitem{rjoub2022explainable}
G.~Rjoub, J.~Bentahar, and O.~A. Wahab, ``{Explainable AI-based Federated Deep Reinforcement Learning for Trusted Autonomous Driving},'' in \emph{2022 International Wireless Communications and Mobile Computing (IWCMC)}.\hskip 1em plus 0.5em minus 0.4em\relax IEEE, 2022, pp. 318--323.

\bibitem{renz2022plant}
\BIBentryALTinterwordspacing
K.~Renz, K.~Chitta, O.-B. Mercea, A.~S. Koepke, Z.~Akata, and A.~Geiger, ``{PlanT: Explainable Planning Transformers via Object-Level Representations},'' in \emph{6th Annual Conference on Robot Learning}, 2022. [Online]. Available: \url{https://openreview.net/forum?id=80vpxjt3vq}
\BIBentrySTDinterwordspacing

\bibitem{hejase2022dynamic}
B.~Hejase, E.~Yurtsever, T.~Han, B.~Singh, D.~P. Filev, H.~E. Tseng, and U.~Ozguner, ``Dynamic and interpretable state representation for deep reinforcement learning in automated driving,'' \emph{IFAC-PapersOnLine}, vol.~55, no.~24, pp. 129--134, 2022.

\bibitem{cultrera2023explaining}
L.~Cultrera, F.~Becattini, L.~Seidenari, P.~Pala, and A.~Del~Bimbo, ``Explaining autonomous driving with visual attention and end-to-end trainable region proposals,'' \emph{Journal of Ambient Intelligence and Humanized Computing}, pp. 1--13, 2023.

\bibitem{shao2023safety}
H.~Shao, L.~Wang, R.~Chen, H.~Li, and Y.~Liu, ``{Safety-Enhanced Autonomous Driving Using Interpretable Sensor Fusion Transformer},'' in \emph{Conference on Robot Learning}.\hskip 1em plus 0.5em minus 0.4em\relax PMLR, 2023, pp. 726--737.

\bibitem{palejalearning}
R.~Paleja, Y.~Niu, A.~Silva, C.~Ritchie, S.~Choi, and M.~Gombolay, ``{Learning Interpretable, High-Performing Policies for Autonomous Driving},'' \emph{Robotics Science and Systems}, 2022.

\bibitem{kenny2023towards}
E.~M. Kenny, M.~Tucker, and J.~Shah, ``{Towards Interpretable Deep Reinforcement Learning with Human-Friendly Prototypes},'' in \emph{The Eleventh International Conference on Learning Representations}, 2023.

\bibitem{yang2023leveraging}
Q.~Yang, H.~Wang, M.~Tong, W.~Shi, G.~Huang, and S.~Song, ``{Leveraging Reward Consistency for Interpretable Feature Discovery in Reinforcement Learning},'' \emph{IEEE Transactions on Systems, Man, and Cybernetics: Systems}, 2023.

\bibitem{lu2024enhancing}
H.~Lu, Y.~Liu, M.~Zhu, C.~Lu, H.~Yang, and Y.~Wang, ``{Enhancing Interpretability of Autonomous Driving Via Human-Like Cognitive Maps: A Case Study on Lane Change},'' \emph{IEEE Transactions on Intelligent Vehicles}, 2024.

\bibitem{liu2024interpretable}
W.~Liu, D.~Li, E.~Aasi, R.~Tron, and C.~Belta, ``Interpretable generative adversarial imitation learning,'' \emph{arXiv preprint arXiv:2402.10310}, 2024.

\bibitem{zhan2019interaction}
W.~Zhan, L.~Sun, D.~Wang, H.~Shi, A.~Clausse, M.~Naumann, J.~Kummerle, H.~Konigshof, C.~Stiller, A.~de~La~Fortelle \emph{et~al.}, ``{INTERACTION dataset: An international, adversarial and cooperative motion dataset in interactive driving scenarios with semantic maps},'' \emph{arXiv preprint arXiv:1910.03088}, 2019.

\bibitem{teng2022hierarchical}
S.~Teng, L.~Chen, Y.~Ai, Y.~Zhou, Z.~Xuanyuan, and X.~Hu, ``{Hierarchical Interpretable Imitation Learning for End-to-End Autonomous Driving},'' \emph{IEEE Transactions on Intelligent Vehicles}, 2022.

\bibitem{PlanT2022}
K.~Renz, K.~Chitta, O.-B. Mercea, A.~S. Koepke, Z.~Akata, and A.~Geiger, ``{PlanT Project Homepage},'' \url{https://www.katrinrenz.de/plant/}, {Accessed online} on February 20, 2024.

\bibitem{omeiza2021towards}
D.~Omeiza, H.~Web, M.~Jirotka, and L.~Kunze, ``{Towards Accountability: Providing Intelligible Explanations in Autonomous Driving},'' in \emph{2021 IEEE Intelligent Vehicles Symposium (IV)}.\hskip 1em plus 0.5em minus 0.4em\relax IEEE, 2021, pp. 231--237.

\bibitem{brewitt2021grit}
C.~Brewitt, B.~Gyevnar, S.~Garcin, and S.~V. Albrecht, ``{GRIT: Fast, Interpretable, and Verifiable Goal Recognition with Learned Decision Trees for Autonomous Driving},'' in \emph{2021 IEEE/RSJ International Conference on Intelligent Robots and Systems (IROS)}.\hskip 1em plus 0.5em minus 0.4em\relax IEEE, 2021, pp. 1023--1030.

\bibitem{cui2022interpretation}
Z.~Cui, M.~Li, Y.~Huang, Y.~Wang, and H.~Chen, ``{An interpretation framework for autonomous vehicles decision-making via SHAP and RF},'' in \emph{2022 6th CAA International Conference on Vehicular Control and Intelligence (CVCI)}.\hskip 1em plus 0.5em minus 0.4em\relax IEEE, 2022, pp. 1--7.

\bibitem{brewitt2023verifiable}
C.~Brewitt, M.~Tamborski, C.~Wang, and S.~V. Albrecht, ``{Verifiable Goal Recognition for Autonomous Driving with Occlusions },'' in \emph{2023 IEEE/RSJ International Conference on Intelligent Robots and Systems (IROS)}.\hskip 1em plus 0.5em minus 0.4em\relax IEEE, 2023, pp. 11\,210--11\,217.

\bibitem{Zhang_2019_CVPR}
Q.~Zhang, Y.~Yang, H.~Ma, and Y.~N. Wu, ``{Interpreting CNNs via Decision Trees},'' in \emph{Proceedings of the IEEE/CVF Conference on Computer Vision and Pattern Recognition (CVPR)}, June 2019.

\bibitem{de2008z3}
L.~De~Moura and N.~Bj{\o}rner, ``{Z3: An efficient SMT solver},'' in \emph{Tools and Algorithms for the Construction and Analysis of Systems: 14th International Conference, TACAS 2008}.\hskip 1em plus 0.5em minus 0.4em\relax Springer, 2008, pp. 337--340.

\bibitem{corso2020interpretable}
A.~Corso and M.~J. Kochenderfer, ``Interpretable safety validation for autonomous vehicles,'' in \emph{2020 IEEE 23rd International Conference on Intelligent Transportation Systems (ITSC)}.\hskip 1em plus 0.5em minus 0.4em\relax IEEE, 2020, pp. 1--6.

\bibitem{suchan2019out}
J.~Suchan, M.~Bhatt, and S.~Varadarajan, ``Out of sight but not out of mind: an answer set programming based online abduction framework for visual sensemaking in autonomous driving,'' in \emph{Proceedings of the 28th International Joint Conference on Artificial Intelligence}, 2019, pp. 1879--1885.

\bibitem{geiger2012we}
A.~Geiger, P.~Lenz, and R.~Urtasun, ``Are we ready for autonomous driving? the kitti vision benchmark suite,'' in \emph{2012 IEEE Conference on Computer Vision and Pattern Recognition}.\hskip 1em plus 0.5em minus 0.4em\relax IEEE, 2012, pp. 3354--3361.

\bibitem{milan2016mot16}
A.~Milan, L.~Leal-Taix{\'e}, I.~Reid, S.~Roth, and K.~Schindler, ``Mot16: A benchmark for multi-object tracking,'' \emph{arXiv preprint arXiv:1603.00831}, 2016.

\bibitem{kothawade2021auto}
S.~Kothawade, V.~Khandelwal, K.~Basu, H.~Wang, and G.~Gupta, ``Auto-discern: Autonomous driving using common sense reasoning,'' \emph{arXiv preprint arXiv:2110.13606}, 2021.

\bibitem{wiegand2019drive}
G.~Wiegand, M.~Schmidmaier, T.~Weber, Y.~Liu, and H.~Hussmann, ``I drive-you trust: Explaining driving behavior of autonomous cars,'' in \emph{{Extended Abstracts of the 2019 CHI Conference on Human Factors in Computing Systems}}, 2019, pp. 1--6.

\bibitem{w2020d}
G.~Wiegand, M.~Eiband, M.~Haubelt, and H.~Hussmann, ``{“I’d like an Explanation for That!” Exploring Reactions to Unexpected Autonomous Driving},'' in \emph{22nd International Conference on Human-Computer Interaction with Mobile Devices and Services}, 2020, pp. 1--11.

\bibitem{decastro2020interpretable}
J.~DeCastro, K.~Leung, N.~Ar{\'e}chiga, and M.~Pavone, ``Interpretable policies from formally-specified temporal properties,'' in \emph{2020 IEEE 23rd International Conference on Intelligent Transportation Systems (ITSC)}.\hskip 1em plus 0.5em minus 0.4em\relax IEEE, 2020, pp. 1--7.

\bibitem{schneider2021explain}
T.~Schneider, J.~Hois, A.~Rosenstein, S.~Ghellal, D.~Theofanou-F{\"u}lbier, and A.~R. Gerlicher, ``{ExplAIn Yourself! Transparency for Positive UX in Autonomous Driving},'' in \emph{Proceedings of the 2021 CHI Conference on Human Factors in Computing Systems}, 2021, pp. 1--12.

\bibitem{schneider2023don}
T.~Schneider, J.~Hois, A.~Rosenstein, S.~Metzl, A.~R. Gerlicher, S.~Ghellal, and S.~Love, ``{Don’t fail me! The Level 5 Autonomous Driving Information Dilemma regarding Transparency and User Experience},'' in \emph{Proceedings of the 28th International Conference on Intelligent User Interfaces}, 2023, pp. 540--552.

\bibitem{kim2023and}
G.~Kim, D.~Yeo, T.~Jo, D.~Rus, and S.~Kim, ``{What and When to Explain? On-road Evaluation of Explanations in Highly Automated Vehicles},'' \emph{Proceedings of the ACM on Interactive, Mobile, Wearable and Ubiquitous Technologies}, vol.~7, no.~3, pp. 1--26, 2023.

\bibitem{radford2018improving}
A.~Radford, K.~Narasimhan, T.~Salimans, and I.~Sutskever, ``{Improving Language Understanding by Generative Pre-Training},'' 2018.

\bibitem{devlin2019bert}
J.~Devlin, M.-W. Chang, K.~Lee, and K.~Toutanova, ``Bert: Pre-training of deep bidirectional transformers for language understanding,'' \emph{Proceedings of the 2019 Conference of the North American Chapter of the Association for Computational Linguistics: Human Language Technologies}, 2019.

\bibitem{brown2020language}
T.~Brown, B.~Mann, N.~Ryder, M.~Subbiah, J.~D. Kaplan, P.~Dhariwal, A.~Neelakantan, P.~Shyam, G.~Sastry, A.~Askell \emph{et~al.}, ``{Language Models are Few-Shot Learners},'' \emph{Advances in Neural Information Processing Systems}, vol.~33, pp. 1877--1901, 2020.

\bibitem{achiam2023gpt}
J.~Achiam, S.~Adler, S.~Agarwal, L.~Ahmad, I.~Akkaya, F.~L. Aleman, D.~Almeida, J.~Altenschmidt, S.~Altman, S.~Anadkat \emph{et~al.}, ``{GPT-4 Technical Report},'' \emph{arXiv preprint arXiv:2303.08774}, 2023.

\bibitem{touvron2023llama}
H.~Touvron, T.~Lavril, G.~Izacard, X.~Martinet, M.-A. Lachaux, T.~Lacroix, B.~Rozi{\`e}re, N.~Goyal, E.~Hambro, F.~Azhar \emph{et~al.}, ``{LLaMA: Open and Efficient Foundation Language Models},'' \emph{arXiv preprint arXiv:2302.13971}, 2023.

\bibitem{touvron2023llama2}
H.~Touvron, L.~Martin, K.~Stone, P.~Albert, A.~Almahairi, Y.~Babaei, N.~Bashlykov, S.~Batra, P.~Bhargava, S.~Bhosale \emph{et~al.}, ``{Llama 2: Open Foundation and Fine-Tuned Chat Models},'' \emph{arXiv preprint arXiv:2307.09288}, 2023.

\bibitem{vicuna2023}
\BIBentryALTinterwordspacing
W.-L. Chiang, Z.~Li, Z.~Lin, Y.~Sheng, Z.~Wu, H.~Zhang, L.~Zheng, S.~Zhuang, Y.~Zhuang, J.~E. Gonzalez, I.~Stoica, and E.~P. Xing, ``{Vicuna: An Open-Source Chatbot Impressing GPT-4 with 90\%* ChatGPT Quality},'' March 2023. [Online]. Available: \url{https://lmsys.org/blog/2023-03-30-vicuna/}
\BIBentrySTDinterwordspacing

\bibitem{taori2023stanford}
R.~Taori, I.~Gulrajani, T.~Zhang, Y.~Dubois, X.~Li, C.~Guestrin, P.~Liang, and T.~B. Hashimoto, ``{Stanford Alpaca: An Instruction-following LLaMA model},'' 2023.

\bibitem{anthropic_claude}
Anthropic, ``{Introducing Claude},'' 2023.

\bibitem{alayrac2022flamingo}
J.-B. Alayrac, J.~Donahue, P.~Luc, A.~Miech, I.~Barr, Y.~Hasson, K.~Lenc, A.~Mensch, K.~Millican, M.~Reynolds \emph{et~al.}, ``{Flamingo: a Visual Language Model for Few-Shot Learning},'' \emph{Advances in Neural Information Processing Systems}, vol.~35, pp. 23\,716--23\,736, 2022.

\bibitem{liu2024visual}
H.~Liu, C.~Li, Q.~Wu, and Y.~J. Lee, ``{Visual Instruction Tuning},'' \emph{Advances in Neural Information Processing Systems}, vol.~36, 2024.

\bibitem{driess2023palm}
D.~Driess, F.~Xia, M.~S. Sajjadi, C.~Lynch, A.~Chowdhery, B.~Ichter, A.~Wahid, J.~Tompson, Q.~Vuong, T.~Yu \emph{et~al.}, ``{PaLM-E: An Embodied Multimodal Language Model},'' in \emph{International Conference on Machine Learning}.\hskip 1em plus 0.5em minus 0.4em\relax PMLR, 2023, pp. 8469--8488.

\bibitem{lin2023video}
B.~Lin, B.~Zhu, Y.~Ye, M.~Ning, P.~Jin, and L.~Yuan, ``{Video-LLaVA: Learning United Visual Representation by Alignment Before Projection},'' \emph{arXiv preprint arXiv:2311.10122}, 2023.

\bibitem{zhang-etal-2023-video}
H.~Zhang, X.~Li, and L.~Bing, ``{"Video-{LL}a{MA}: An Instruction-tuned Audio-Visual Language Model for Video Understanding"},'' in \emph{Proceedings of the 2023 Conference on Empirical Methods in Natural Language Processing: System Demonstrations}, 2023, pp. 543--553.

\bibitem{team2023gemini}
R.~Anil, S.~Borgeaud, Y.~Wu, J.-B. Alayrac, J.~Yu, R.~Soricut, J.~Schalkwyk, A.~M. Dai, A.~Hauth \emph{et~al.}, ``{Gemini: A Family of Highly Capable Multimodal Models},'' \emph{arXiv preprint arXiv:2312.11805}, 2023.

\bibitem{anthropic_claude3}
Anthropic, ``{Introducing the next generation of Claude},'' https://www.anthropic.com/news/claude-3-family, 2024.

\bibitem{wayve_lingo1_2023}
{Waywe Research}, ``{LINGO-1: Exploring Natural Language for Autonomous Driving},'' https://wayve.ai/thinking/lingo-natural-language-autonomous-driving/, 2023.

\bibitem{wayve_lingo2_2024}
{Waywe Research Team}, ``{LINGO-2: Driving with Natural Language},'' https://wayve.ai/thinking/lingo-2-driving-with-language/, 2024.

\bibitem{chen2023driving}
L.~Chen, O.~Sinavski, J.~H{\"u}nermann, A.~Karnsund, A.~J. Willmott, D.~Birch, D.~Maund, and J.~Shotton, ``{Driving with LLMs: Fusing Object-Level Vector Modality for Explainable Autonomous Driving},'' \emph{arXiv preprint arXiv:2310.01957}, 2023.

\bibitem{xu2023drivegpt4}
Z.~Xu, Y.~Zhang, E.~Xie, Z.~Zhao, Y.~Guo, K.~K. Wong, Z.~Li, and H.~Zhao, ``{DriveGPT4: Interpretable End-to-End Autonomous Driving via Large Language Model},'' \emph{arXiv preprint arXiv:2310.01412}, 2023.

\bibitem{marcu2023lingoqa}
A.-M. Marcu, L.~Chen, J.~H{\"u}nermann, A.~Karnsund, B.~Hanotte, P.~Chidananda, S.~Nair, V.~Badrinarayanan, A.~Kendall, J.~Shotton \emph{et~al.}, ``{LingoQA: Video Question Answering for Autonomous Driving},'' \emph{arXiv preprint arXiv:2312.14115}, 2023.

\bibitem{park2024vlaad}
S.~Park, M.~Lee, J.~Kang, H.~Choi, Y.~Park, J.~Cho, A.~Lee, and D.~Kim, ``{VLAAD: Vision and Language Assistant for Autonomous Driving},'' in \emph{Proceedings of the IEEE/CVF Winter Conference on Applications of Computer Vision}, 2024, pp. 980--987.

\bibitem{mao2023gpt}
J.~Mao, Y.~Qian, H.~Zhao, and Y.~Wang, ``{GPT-Driver: Learning to Drive with GPT },'' \emph{NeurIPS 2023 Foundation Models for Decision Making Workshop}, 2023.

\bibitem{sha2023languagempc}
H.~Sha, Y.~Mu, Y.~Jiang, L.~Chen, C.~Xu, P.~Luo, S.~E. Li, M.~Tomizuka, W.~Zhan, and M.~Ding, ``{LanguageMPC: Large Language Models as Decision Makers for Autonomous Driving},'' \emph{arXiv preprint arXiv:2310.03026}, 2023.

\bibitem{wen2023dilu}
L.~Wen, D.~Fu, X.~Li, X.~Cai, T.~Ma, P.~Cai, M.~Dou, B.~Shi, L.~He, and Y.~Qiao, ``{DiLu: A Knowledge-Driven Approach to Autonomous Driving with Large Language Models},'' \emph{International Conference on Learning Representations}, 2024.

\bibitem{yuan2024rag}
J.~Yuan, S.~Sun, D.~Omeiza, B.~Zhao, P.~Newman, L.~Kunze, and M.~Gadd, ``Rag-driver: Generalisable driving explanations with retrieval-augmented in-context learning in multi-modal large language model,'' \emph{arXiv preprint arXiv:2402.10828}, 2024.

\bibitem{dewangan2023talk2bev}
V.~Dewangan, T.~Choudhary, S.~Chandhok, S.~Priyadarshan, A.~Jain, A.~K. Singh, S.~Srivastava, K.~M. Jatavallabhula, and K.~M. Krishna, ``{Talk2BEV: Language-enhanced Bird's-eye View Maps for Autonomous Driving},'' \emph{arXiv preprint arXiv:2310.02251}, 2023.

\bibitem{atakishiyev2024safety}
S.~Atakishiyev, M.~Salameh, and R.~Goebel, ``{Safety Implications of Explainable Artificial Intelligence in End-to-End Autonomous Driving},'' \emph{arXiv preprint arXiv:2403.12176}, 2024.

\bibitem{shen2022to}
Y.~Shen, S.~Jiang, Y.~Chen, and K.~R. Driggs-Campbell, ``{To Explain or Not to Explain: A Study on the Necessity of Explanations for Autonomous Vehicles},'' in \emph{{NeurIPS Workshop on Progress and Challenges in Building Trustworthy Embodied AI}}, 2022.

\bibitem{atakishiyev2023towards}
S.~Atakishiyev, M.~Salameh, H.~Yao, and R.~Goebel, ``{Towards Safe, Explainable, and Regulated Autonomous Driving},'' \emph{Explainable Artificial Intelligence for Intelligent Transportation Systems}, pp. 32--52, 2023.

\bibitem{ISO26262_6}
\BIBentryALTinterwordspacing
{ISO 26262-6}. {{ISO 26262-6:2018 Road vehicles — Functional safety — Part 6: Product development at the software level}}. (Accessed on February 10, 2023). [Online]. Available: \url{https://www.iso.org/standard/68388.html}
\BIBentrySTDinterwordspacing

\bibitem{johansen2016ship}
T.~A. Johansen, T.~Perez, and A.~Cristofaro, ``{Ship Collision Avoidance and COLREGS Compliance Using Simulation-Based Control Behavior Selection With Predictive Hazard Assessment},'' \emph{IEEE Transactions on Intelligent Transportation Systems}, vol.~17, no.~12, pp. 3407--3422, 2016.

\bibitem{fu2023drive}
D.~Fu, X.~Li, L.~Wen, M.~Dou, P.~Cai, B.~Shi, and Y.~Qiao, ``Drive like a human: Rethinking autonomous driving with large language models,'' \emph{arXiv preprint arXiv:2307.07162}, 2023.

\bibitem{nie2023reason2drive}
M.~Nie, R.~Peng, C.~Wang, X.~Cai, J.~Han, H.~Xu, and L.~Zhang, ``Reason2drive: Towards interpretable and chain-based reasoning for autonomous driving,'' \emph{arXiv preprint arXiv:2312.03661}, 2023.

\bibitem{chi2024multi}
F.~Chi, Y.~Wang, P.~Nasiopoulos, and V.~C. Leung, ``Multi-modal gpt-4 aided action planning and reasoning for self-driving vehicles,'' in \emph{ICASSP 2024-2024 IEEE International Conference on Acoustics, Speech and Signal Processing (ICASSP)}.\hskip 1em plus 0.5em minus 0.4em\relax IEEE, 2024, pp. 7325--7329.

\bibitem{duan2024prompting}
Y.~Duan, Q.~Zhang, and R.~Xu, ``Prompting multi-modal tokens to enhance end-to-end autonomous driving imitation learning with llms,'' \emph{arXiv preprint arXiv:2404.04869}, 2024.

\bibitem{softwarecorrectness1976}
S.~Hantler and J.~King, ``{An Introduction to Proving the Correctness of Programs},'' \emph{ACM Computing Surveys}, vol.~8, pp. 331--353, 09 1976.

\bibitem{ISO_cybersecurity}
\BIBentryALTinterwordspacing
{ISO Technical Commitee}, ``{ISO/SAE 21434:2021 Road vehicles — Cybersecurity engineering},'' 2021. [Online]. Available: \url{https://www.iso.org/standard/70918.html}
\BIBentrySTDinterwordspacing

\bibitem{qayyum2020securing}
A.~Qayyum, M.~Usama, J.~Qadir, and A.~Al-Fuqaha, ``Securing connected \& autonomous vehicles: Challenges posed by adversarial machine learning and the way forward,'' \emph{IEEE Communications Surveys \& Tutorials}, vol.~22, no.~2, pp. 998--1026, 2020.

\bibitem{kim2021cybersecurity}
K.~Kim, J.~S. Kim, S.~Jeong, J.-H. Park, and H.~K. Kim, ``Cybersecurity for autonomous vehicles: Review of attacks and defense,'' \emph{Computers \& security}, vol. 103, p. 102150, 2021.

\bibitem{sun2021survey}
X.~Sun, F.~R. Yu, and P.~Zhang, ``{A Survey on Cyber-Security of Connected and Autonomous Vehicles (CAVs)},'' \emph{IEEE Transactions on Intelligent Transportation Systems}, vol.~23, no.~7, pp. 6240--6259, 2021.

\bibitem{koo2015did}
J.~Koo, J.~Kwac, W.~Ju, M.~Steinert, L.~Leifer, and C.~Nass, ``Why did my car just do that? explaining semi-autonomous driving actions to improve driver understanding, trust, and performance,'' \emph{International Journal on Interactive Design and Manufacturing (IJIDeM)}, vol.~9, no.~4, pp. 269--275, 2015.

\bibitem{haspiel2018explanations}
J.~Haspiel, N.~Du, J.~Meyerson, L.~P. Robert~Jr, D.~Tilbury, X.~J. Yang, and A.~K. Pradhan, ``Explanations and expectations: Trust building in automated vehicles,'' in \emph{Companion of the 2018 ACM/IEEE international conference on human-robot interaction}, 2018, pp. 119--120.

\bibitem{huang2022takeover}
G.~Huang and B.~J. Pitts, ``Takeover requests for automated driving: The effects of signal direction, lead time, and modality on takeover performance,'' \emph{Accident Analysis \& Prevention}, vol. 165, p. 106534, 2022.

\bibitem{mok2015emergency}
B.~Mok, M.~Johns, K.~J. Lee, D.~Miller, D.~Sirkin, P.~Ive, and W.~Ju, ``{Emergency, Automation Off: Unstructured Transition Timing for Distracted Drivers of Automated Vehicles},'' in \emph{2015 IEEE 18th International Conference on Intelligent Transportation Systems}.\hskip 1em plus 0.5em minus 0.4em\relax IEEE, 2015, pp. 2458--2464.

\bibitem{wan2018effects}
J.~Wan and C.~Wu, ``The effects of lead time of take-over request and nondriving tasks on taking-over control of automated vehicles,'' \emph{IEEE Transactions on Human-Machine Systems}, vol.~48, no.~6, pp. 582--591, 2018.

\bibitem{schieben2019designing}
A.~Schieben, M.~Wilbrink, C.~Kettwich, R.~Madigan, T.~Louw, and N.~Merat, ``Designing the interaction of automated vehicles with other traffic participants: design considerations based on human needs and expectations,'' \emph{Cognition, Technology \& Work}, vol.~21, pp. 69--85, 2019.

\bibitem{thrun2006stanley}
S.~Thrun, M.~Montemerlo, H.~Dahlkamp, D.~Stavens, A.~Aron, J.~Diebel, P.~Fong, J.~Gale, M.~Halpenny, G.~Hoffmann \emph{et~al.}, ``{Stanley: The robot that won the DARPA Grand Challenge},'' \emph{Journal of Field Robotics}, vol.~23, no.~9, pp. 661--692, 2006.

\bibitem{jain2021autonomy}
A.~Jain, L.~Del~Pero, H.~Grimmett, and P.~Ondruska, ``Autonomy 2.0: Why is self-driving always 5 years away?'' \emph{arXiv preprint arXiv:2107.08142}, 2021.

\bibitem{hawke2021reimagining}
J.~Hawke, V.~Badrinarayanan, A.~Kendall \emph{et~al.}, ``Reimagining an autonomous vehicle,'' \emph{arXiv preprint arXiv:2108.05805}, 2021.

\bibitem{Waymo_e2e_2024}
\BIBentryALTinterwordspacing
{Erez Dagan}, ``{Solving the long-tail with e2e AI: “The revolution will not be supervised”},'' 2024. [Online]. Available: \url{https://wayve.ai/thinking/e2e-embodied-ai-solves-the-long-tail/}
\BIBentrySTDinterwordspacing

\bibitem{duan2022survey}
J.~Duan, S.~Yu, H.~L. Tan, H.~Zhu, and C.~Tan, ``{A Survey of Embodied AI: From Simulators to Research Tasks},'' \emph{IEEE Transactions on Emerging Topics in Computational Intelligence}, vol.~6, no.~2, pp. 230--244, 2022.

\bibitem{wang2024comprehensive}
L.~Wang, X.~Zhang, H.~Su, and J.~Zhu, ``{A Comprehensive Survey of Continual Learning: Theory, Method and Application},'' \emph{IEEE Transactions on Pattern Analysis and Machine Intelligence}, 2024.

\bibitem{amodei2016concrete}
D.~Amodei, C.~Olah, J.~Steinhardt, P.~Christiano, J.~Schulman, and D.~Man{\'e}, ``{Concrete problems in AI safety},'' \emph{arXiv preprint arXiv:1606.06565}, 2016.

\bibitem{waymo_driver2024}
{Waymo Team}, ``{Self-Driving Car Technology for a Reliable Ride},'' 2024.

\bibitem{hadfield2016cooperative}
D.~Hadfield-Menell, S.~J. Russell, P.~Abbeel, and A.~Dragan, ``{Cooperative Inverse Reinforcement Learning},'' \emph{Advances in Neural Information Processing Systems}, vol.~29, 2016.

\bibitem{toghi2021cooperative}
B.~Toghi, R.~Valiente, D.~Sadigh, R.~Pedarsani, and Y.~P. Fallah, ``{Cooperative Autonomous Vehicles that Sympathize with Human Drivers},'' in \emph{2021 IEEE/RSJ International Conference on Intelligent Robots and Systems (IROS)}.\hskip 1em plus 0.5em minus 0.4em\relax IEEE, 2021, pp. 4517--4524.

\bibitem{knox2023reward}
W.~B. Knox, A.~Allievi, H.~Banzhaf, F.~Schmitt, and P.~Stone, ``Reward (mis) design for autonomous driving,'' \emph{Artificial Intelligence}, vol. 316, p. 103829, 2023.

\bibitem{molly2020}
\BIBentryALTinterwordspacing
{ITU Team}, ``{The Molly Problem},'' \emph{AI for autonomous and assisted driving}, (Accessed on February 12, 2024). [Online]. Available: \url{https://www.itu.int/en/ITU-T/focusgroups/ai4ad/Pages/MollyProblem.aspx}
\BIBentrySTDinterwordspacing

\bibitem{filos2020can}
A.~Filos, P.~Tigkas, R.~McAllister, N.~Rhinehart, S.~Levine, and Y.~Gal, ``{Can Autonomous Vehicles Identify, Recover From, and Adapt to Distribution Shifts?}'' in \emph{International Conference on Machine Learning}.\hskip 1em plus 0.5em minus 0.4em\relax PMLR, 2020, pp. 3145--3153.

\bibitem{magdici2016fail}
S.~Magdici and M.~Althoff, ``{Fail-Safe Motion Planning of Autonomous Vehicles},'' in \emph{2016 IEEE 19th International Conference on Intelligent Transportation Systems (ITSC)}.\hskip 1em plus 0.5em minus 0.4em\relax IEEE, 2016, pp. 452--458.

\bibitem{xue2023fail}
W.~Xue, Z.~Wang, R.~Zheng, X.~Mei, B.~Yang, and K.~Nakano, ``{Fail-Safe Behavior and Motion Planning Incorporating Shared Control for Potential Driver Intervention},'' \emph{IEEE Transactions on Intelligent Vehicles}, 2023.

\bibitem{pek2020fail}
C.~Pek and M.~Althoff, ``{Fail-Safe Motion Planning for Online Verification of Autonomous Vehicles Using Convex Optimization},'' \emph{IEEE Transactions on Robotics}, vol.~37, no.~3, pp. 798--814, 2020.

\bibitem{kenny2023pursuit}
E.~Kenny and J.~Shah, ``{In Pursuit of Regulatable LLMs},'' in \emph{NeurIPS 2023 Workshop on Regulatable ML}, 2023.

\bibitem{sanneman2022situation}
L.~Sanneman and J.~A. Shah, ``{The Situation Awareness Framework for Explainable AI (SAFE-AI) and Human Factors Considerations for XAI Systems},'' \emph{International Journal of Human--Computer Interaction}, vol.~38, no. 18-20, pp. 1772--1788, 2022.

\bibitem{endsley2023supporting}
M.~R. Endsley, ``{Supporting Human-AI Teams: Transparency, explainability, and situation awareness},'' \emph{Computers in Human Behavior}, vol. 140, p. 107574, 2023.

\bibitem{diaz2023connecting}
N.~D{\'\i}az-Rodr{\'\i}guez, J.~Del~Ser, M.~Coeckelbergh, M.~L. de~Prado, E.~Herrera-Viedma, and F.~Herrera, ``{Connecting the dots in trustworthy Artificial Intelligence: From AI principles, ethics, and key requirements to responsible AI systems and regulation},'' \emph{Information Fusion}, vol.~99, p. 101896, 2023.

\bibitem{zhang2024critical}
T.~Zhang, W.~Li, W.~Huang, and L.~Ma, ``Critical roles of explainability in shaping perception, trust, and acceptance of autonomous vehicles,'' \emph{International Journal of Industrial Ergonomics}, vol. 100, p. 103568, 2024.

\bibitem{ehsan2021operationalizing}
U.~Ehsan, P.~Wintersberger, Q.~V. Liao, M.~Mara, M.~Streit, S.~Wachter, A.~Riener, and M.~O. Riedl, ``{Operationalizing Human-Centered Perspectives in Explainable AI},'' in \emph{Extended Abstracts of the 2021 CHI Conference on Human Factors in Computing Systems}, 2021, pp. 1--6.

\end{thebibliography}
\bibliographystyle{IEEEtran}

\vspace{-0.5cm}
\begin{IEEEbiography}
    [{\includegraphics[width=1in,height=1.25in,clip, keepaspectratio]{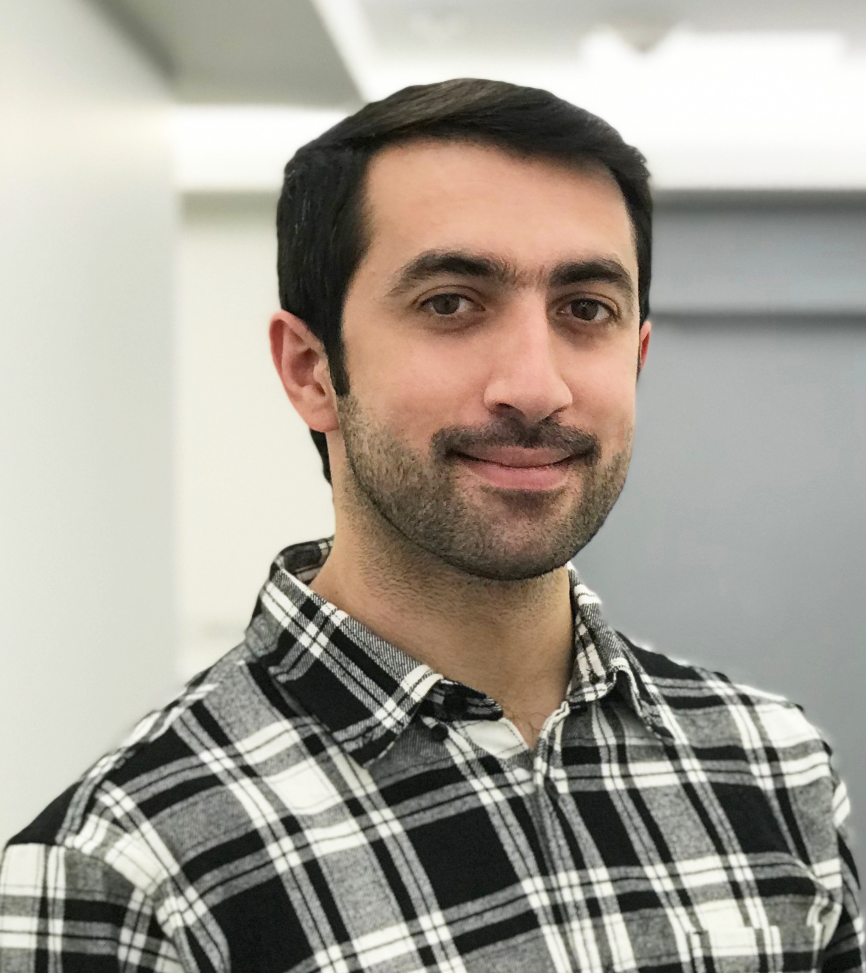}}]{Shahin Atakishiyev}
is from the Mirzabayli village of Gabala, Azerbaijan. He received a BSc Computer Engineering degree from Qafqaz University, Azerbaijan in June 2015, and an MSc Computer Engineering degree with a specialization in Software Engineering and Intelligent Systems from the University of Alberta, Canada in January 2018. Currently, he is a PhD Candidate in the Department of Computing Science at the University of Alberta, working at the Explainable Artificial Intelligence (XAI) lab under the supervision of Prof. Randy Goebel. Shahin’s research interests include Safe, Ethical, and Explainable Artificial Intelligence, and its application to real-world problems.
\end{IEEEbiography}
\vspace{-0.5cm}
\begin{IEEEbiography}
    [{\includegraphics[width=1in,clip, keepaspectratio]{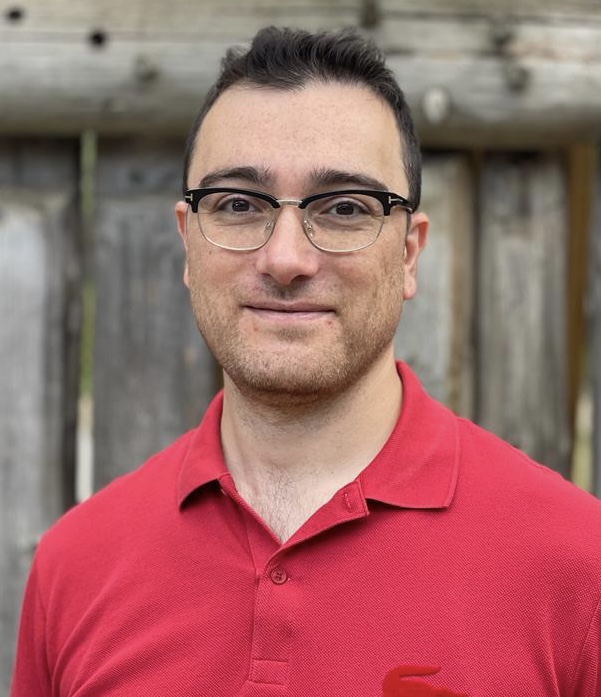}}]{Mohammad Salameh}
received the Ph.D. degree from the University of Alberta under the supervision of Dr. Greg Kondrak and Dr. Colin Cherry, with the main focus on statistical machine translation and sentiment analysis. He is currently a Principal Researcher at Huawei Technologies Canada Company Ltd and leading the neural architecture search group, focusing on gradient-based and reinforcement learning approaches. He co-organized Determining Sentiment Intensity in Tweets (SemEval2016) and Affects in Tweets (SemEval2018) shared tasks.
\end{IEEEbiography}

\begin{IEEEbiography}[{\includegraphics[width=1in,clip, keepaspectratio]{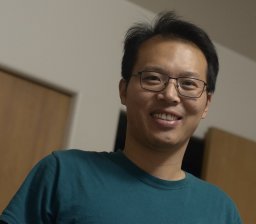}}]{Hengshuai Yao}
received the Ph.D. degree in reinforcement learning from the Reinforcement Learning and Artificial Intelligence (RLAI) Laboratory, Department of Computing Science, University of Alberta, in 2014. His thesis is on model-based reinforcement learning with linear function approximation. During his Ph.D. studies, he worked on reinforcement learning theory, algorithms, and web applications. He joined NCSoft Game Studio, San Francisco, in 2016, where he worked on reinforcement learning in games. Yao is currently working as a Senior Research Scientist at Sony AI.
\end{IEEEbiography}

\begin{IEEEbiography}
[{\includegraphics[width=1in,clip, keepaspectratio]{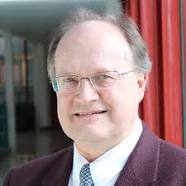}}]{R.G. (Randy) Goebel}
 is currently a Professor of Computing Science in the Department of Computing Science at the University of Alberta and Fellow and co-founder of the Alberta Machine Intelligence Institute (Amii). He received the B.Sc. (Computer Science), M.Sc. (Computing Science), and Ph.D. (Computer Science) from the Universities of Regina, Alberta, and British Columbia, respectively.\\
Professor Goebel's theoretical work on abduction, hypothetical reasoning and belief revision is internationally well known; his recent research is focused on the formalization of visualization and explainable artificial intelligence (XAI), especially in applications in autonomous driving, legal reasoning, and precision health. He has worked on optimization, algorithm complexity, systems biology, natural language processing, and automated reasoning.

Randy has previously held faculty appointments at the University of Waterloo, University of Tokyo, Multimedia University (Kuala Lumpur), Hokkaido University (Sapporo), visiting researcher engagements at National Institute of Informatics (Tokyo), DFKI (Germany), and NICTA (now Data61, Australia); he is actively involved in collaborative research projects in Canada, Japan, Germany, France, the UK, and China.
\end{IEEEbiography}
\end{document}